\documentclass[pdflatex,sn-basic,Numbered]{sn-jnl}

\usepackage{graphicx}%
\usepackage{multirow}%
\usepackage{amsmath,amssymb,amsfonts}%
\usepackage{amsthm}%
\usepackage{mathrsfs}%
\usepackage[title]{appendix}%
\usepackage{xcolor}%
\usepackage{textcomp}%
\usepackage{manyfoot}%
\usepackage{booktabs}%
\usepackage{algorithm}%
\usepackage{algorithmicx}%
\usepackage{algpseudocode}%
\usepackage{listings}%
\usepackage{graphicx}%
\usepackage{multirow}%
\usepackage{xspace}%
\usepackage{hyperref}%
\usepackage[nameinlink]{cleveref}%
\usepackage{pdfpages}%

\theoremstyle{thmstyleone}%
\theoremstyle{thmstyletwo}%

\theoremstyle{thmstylethree}%

\newcommand{\kilogram}{\textsc{KiloGram}\xspace}
\newcommand{\aautoref}[1]{\hyperref[#1]{Appendix~\ref*{#1}}}

\begin{document}

\title[Ad hoc conventions generalize to new referents]{Ad hoc conventions generalize to new referents}


\author[1]{\fnm{Anya} \sur{Ji}}\email{anyaji@berkeley.edu}

\author[2]{\fnm{Claire Augusta} \sur{Bergey}}\email{cbergey@stanford.edu}

\author[3]{\fnm{Ron} \sur{Eliav}}\email{roneliav1@gmail.com}

\author[4]{\fnm{Yoav} \sur{Artzi}}\email{yoav@cs.cornell.edu}

\author*[2]{\fnm{Robert D.} \sur{Hawkins}}\email{rdhawkins@stanford.edu}

\affil[1]{\orgdiv{Department of Electrical Engineering and Computer Sciences}, \orgname{University of California, Berkeley}}

\affil*[2]{\orgdiv{Department of Linguistics}, \orgname{Stanford University}}

\affil[3]{\orgdiv{Department of Computer Science}, \orgname{Bar-Ilan University}}

\affil[4]{\orgdiv{Department of Computer Science}, \orgname{Cornell University}}


\abstract{How do people talk about things they've never talked about before?
One view suggests that a new shared naming system establishes an arbitrary link to a specific target, like proper names that cannot extend beyond their bearers. 
An alternative view proposes that forming a shared way of describing objects involves broader \emph{conceptual alignment}, reshaping each individual's semantic space in ways that should generalize to new referents.
We test these competing accounts in a dyadic communication study (N=302) leveraging the recently-released \kilogram dataset containing over 1,000 abstract tangram images.
After pairs of participants coordinated on referential conventions for one set of images through repeated communication, we measured the extent to which their descriptions aligned for \emph{undiscussed images}.
We found strong evidence for generalization: partners showed increased alignment relative to their pre-test labels. Generalization also decayed nonlinearly with visual similarity (consistent with Shepard's law) and was robust across levels of the images' nameability.
These findings suggest that ad hoc conventions are not arbitrary labels but reflect genuine conceptual coordination, with implications for theories of reference and the design of more adaptive language agents.}




\maketitle


\section{Introduction}\label{sec1}

Humans have a remarkable ability to communicate about novel experiences, objects, and ideas.
When we need to talk about something we have never talked about before, whether we are gazing up at an oddly shaped cloud or labeling a condition in a complex experimental design \cite{barsalou1983ad,CasasantoLupyan15}, we must solve a challenging social coordination problem \cite{Schelling1960,Lewis1969,sperber1986relevance,levinson2025interaction,young1993evolution}.
Through a rapid process of negotiation and adaptation --- offering possible descriptions, receiving feedback, adjusting, clarifying, and refining --- two people can establish mutually-understandable referring expressions even when they weren't initially able to understand one another \cite{clark1996using,brennan1996conceptual,garrod1987saying}. 
These \emph{ad hoc conventions} arise from the practical demands of communication without external instruction or dictionary definitions \cite{hawkins2022partners}, raising a fundamental question about the nature of linguistic innovation:
When people create a way to refer to unfamiliar entities, what exactly are they creating?

One possibility is that newly minted referential conventions function much like someone's name: they create an arbitrary link between a string of words and a single unique entity \cite{mill1system,carroll1980naming}. 
Under the strongest form of a \emph{rigid designator} view, each entity requires its own dedicated ``baptism''  \cite{kripke1980naming} and conventions remain tethered to the particular entity for which they arose. 
Alternatively, the process of successful coordination might involve something more wide-reaching: a mutual reshaping of each person's conceptual space that affects not just how partners talk about the original target, but how they understand and describe related entities they encounter later \cite{stolk2013neural,stolk2016conceptual,rasenberg2020alignment}.
This \emph{conceptual alignment} account draws on research suggesting that concepts are organized in high-dimensional representational spaces in which proximity reflects psychological similarity \cite{nosofsky1986attention,shepard1987toward,gardenfors2000conceptual,rogers2004semantic,landauer1997solution}, such that an ad hoc convention can readily generalize to nearby referents. 
The distinction between these accounts has implications in further-reaching fields: results could inform the design of more adaptive AI systems that learn to coordinate with human partners in new environments \cite{du2024constrained,dafoe2020open}, and shed light on how children acquire the productive capacity to talk about novel experiences through interaction \cite{clark2022language,tomasello2005constructing}.

Relevant evidence comes from repeated reference games using deliberately ambiguous referential targets, such as abstract drawings or tangram shapes~\citep{krauss1964changes,clark1986referring,hawkins2020characterizing}.
A key finding from these studies is that participants typically begin with verbose indefinite descriptions appealing to multiple features of the target (e.g. ``a dog smiling and looking left with its tail pointed up''), but gradually converge on a set of short and idiosyncratic names (e.g. ``happy dog''). 
These ad hoc conventions are relatively sticky, as speakers will keep using the same name even in new referential contexts where the necessary degree of specificity has been relaxed~\citep{brennan1996conceptual,ibarra2016flexibility}, and also partner-specific, as speakers do not necessarily extend the same label when talking with a new partner~\citep{wilkes1992coordinating,metzing2003conceptual,brown2015people}.
When the task goals shift, participants flexibly abandon previously negotiated names without explicit renegotiation, and adopt new referring expressions that better fit the changed context \cite{ibarra2016flexibility, brennan1996conceptual}. 
Recent work has provided increasingly sophisticated evidence for the alignment process \cite{ghaleb2024analysing}, using automated detection of ``shared constructions'' in referential communication to show that speakers not only converge on referring expressions during an interaction, but that their private post-interaction labels for the same entities were also strongly aligned.

\begin{figure*}[t]
    \centering
    \includegraphics[keepaspectratio=true, width=\textwidth]{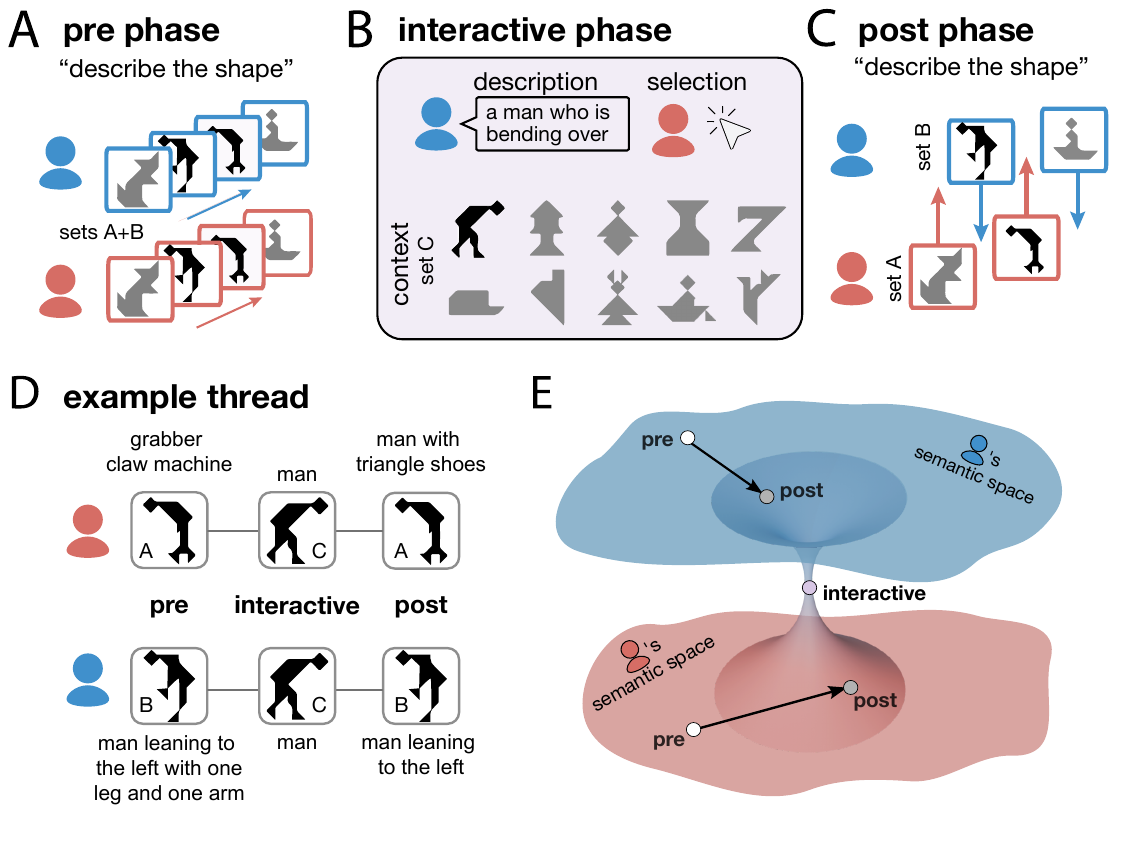}
\caption{We examined how pairs of participants aligned their representations of novel tangram shapes across three phases. (A-C) Participants independently described tangrams in a \emph{pre} phase, then played a repeated reference game with a different set of shapes in an \emph{interactive} phase, and finally described the original shapes again in a \emph{post} phase. Highlighted tangrams (colored black) are in the same thread.
(D) Tangrams were organized into ``threads'' of three visually similar shapes, with one from each context (set A, B, and C). In this example, tangrams A, B, and C constitute a thread, appearing across all three phases for a given pair of participants. Example descriptions in each phase are presented above/below the corresponding tangram. (E) We tested whether alignment on shared conventions during interaction (e.g., ``man'') would generalize to nearby undiscussed shapes, with generalization strength decreasing nonlinearly with visual distance, consistent with Shepard's law. The diagram illustrates hypothesized convergence from pre (white dots) to post (gray dots) for both players.}
 \label{fig:visualization}
\end{figure*}

These findings indicate flexibility in referring to the \emph{same} entities across contexts.
However, a critical axis of generalization has been comparatively under-examined: the problem of generalization to \emph{new entities}~\citep{shepard1987toward,nolle2018emergence,raviv2022variability}.
If interlocutors are not simply memorizing a one-to-one rigid designator but achieving some form of broader \emph{conceptual alignment}~\citep{rane2024concept,maiorca2024latent,clark1996using,stolk2016conceptual,sucholutsky2023getting,garrod1987saying,poncelet2025fine}, then we should observe a gradient of generalization to nearby targets in the conceptual space.
This prediction aligns with broader principles of generalization observed across cognitive domains, from categorization \cite{nosofsky1986attention,ashby2005human} to concept learning \cite{tenenbaum2001generalization} and motor learning \cite{poh2021generalization}. 
Shepard's Universal Law suggests that if referential conventions engage general learning mechanisms, they should exhibit similar distance-dependent generalization gradients \cite{shepard1987toward,marjieh2024universal,chater2003generalized}.

A major methodological obstacle to quantifying referential generalization has been the absence of diverse stimuli with granular variation and reliable methods for measuring psychological distances between abstract stimuli like tangrams. Traditional reference game studies have relied on small sets of targets, making it difficult to systematically explore generalization gradients across a wide range of visual similarities.
Recent advances now make it possible to address these limitations systematically.
We leverage the \kilogram dataset~\citep{Ji2022AbstractVR}, which provides orders of magnitude more stimuli than previously available, enabling fine-grained exploration of the visual similarity space. We combine this with state-of-the-art vision and language models to compute independent measures of visual and textual similarity that align with human perceptual judgments~\cite{fan2018common,peterson2018evaluating,jha2023extracting}. 

These methodological advances allow us to design a pre-post reference game experiment in which pairs of participants coordinate on conventions for one set of tangrams and later describe visually similar tangrams they never discussed together. This design directly pits the competing theoretical accounts against one another: if conventions function as rigid designators, we should observe no increased alignment for undiscussed tangrams regardless of their visual similarity to discussed targets. If conventions reflect conceptual alignment, we should observe graded generalization that falls off systematically with visual distance from the discussed targets.


\section{Results}\label{sec2}

\begin{figure*}[!t]
\centering
\includegraphics[width=\textwidth]{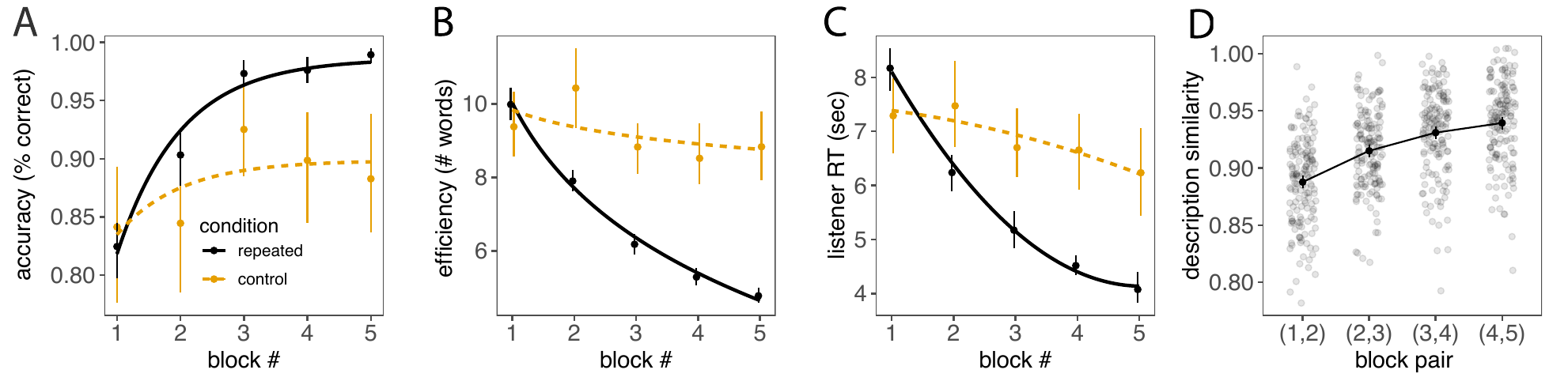}
\caption{Participants' performance improved over time for the repeated set of targets, relative to a control set, in terms of their (A) communicative accuracy (proportion of trials in which the listener was able to correctly select the target), (B) communicative efficiency (the number of words in the utterances sent by the speaker and the listener), and (C) speed (the time elapsed between the first speaker message and the listener selection). 
Additionally, (D) participants increasingly aligned on the same descriptions; similarity between the descriptions used on successive blocks is measured by cosine similarity using text-embedding-ada-002 embeddings \cite{openaiEmbeddings}. Error bars are bootstrapped 95\% CIs and curves are best-fitting quadratic regression fits.}
\label{fig:stats}
\end{figure*}

\subsection{Participants successfully establish referential conventions}

We recruited 163 pairs of participants to participate in a three-stage communication game (\autoref{fig:visualization}A-C). 12 pairs with incomplete games were excluded, leaving 151 ($N=302$) pairs for analysis.
In the pre-interactive phase (abbreviated as the \textbf{pre phase}), we asked each participant to independently type descriptions of a series of 20 tangrams, abstract figures composed of basic geometric shapes.
Then, in the \textbf{interactive phase}, participants played an interactive reference game with a set of 10 unseen tangrams that varied in their similarity to the initial set (\autoref{fig:visualization}D). 
In this phase, they were assigned to ``speaker'' and ``listener'' roles: the speaker was asked to describe a target tangram such that the listener could accurately choose it out of a grid of the 10 possible choices. 
This phase consisted of five blocks, with six trials per block. 
Five tangrams appeared as the target on every block (repeated condition), while the other five were interspersed such that they appeared as target only once (control condition). 
Participants switched roles each block and received full feedback (i.e. the speaker saw the listener's selection, and the listener saw the intended target referent).
Finally, in the post-interactive phase (abbreviated as the \textbf{post phase}), we asked participants to describe the same 20 tangrams they saw in the pre phase for one another, with no feedback about accuracy or opportunity for clarification. 
This pre-post design allowed us to test whether conventions formed during interaction would generalize to a held-out set of tangrams.

We begin by establishing that participants successfully formed conventions during the interactive phase. 
To isolate item-specific learning from general task improvement (e.g., fatigue or familiarity with the task), we compared performance on repeated versus control targets. 
To the extent that changes in the referring expressions for the repeated targets are due to generic task pressures, both conditions should show similar gains.
However, if participants formed target-specific conventions, we should see selectively stronger improvements for repeated items, giving us confidence that any pre-to-post generalization reflects genuine convention formation rather than task artifacts.

We examined three performance metrics to assess the convention formation process: the accuracy with which the listener was able to choose the target, the length of the speaker and the listener's chat in words, and the speed of the listener's referent selection. 
For each metric, we built a multi-level Bayesian regression model using the \texttt{brms} package \cite{burkner2017brms}, with fixed effects for repetition block (integers 1 to 5) and condition (repeated vs. control) as well as their interaction.
To account for nonlinearities, we also included a quadratic effect of repetition block.
We included a maximal random effect structure at the dyad level, with random intercepts, slopes, and interaction.

Beginning with accuracy, we used a Bernoulli linking function to predict trial-level response accuracy and found a strong simple effect of repetition block for the repeated condition ($b = 74.7$, 95\% credible interval: $[60.4, 90.3]$), demonstrating that dyads communicate more accurately about repeated targets over successive repetitions. 
We also found a significant interaction of block with condition $(b = -62.4$, 95\% credible interval: $[-83.9, -41.6]$), demonstrating that these improvements were item-specific: they were specific to tangrams about which partners had a shared communicative history, and did not accrue to tangrams merely as a result of appearing later in the session (\autoref{fig:stats}A). 

To investigate whether partners became more communicatively efficient, we examined the length of dyads' communication and the listener's response time to select the corresponding image. Predicting length of dyads' exchange on each trial using the same model structure as above and a Poisson linking function, we found linear and quadratic effects of block in the repeated condition such that speakers use fewer words to refer as the experiment progresses ($b = -17.41$, 95\% credible interval: $[-19.25, -15.68]$) and their efficiency drops more quickly in earlier blocks (quadratic effect: $b = 2.40$, 95\% credible interval: $[1.24, 3.53]$) (\autoref{fig:stats}B), consistent with classic observations~\citep{krauss1964changes,clark1986referring}. We also found interactions between repetition condition and both simple and quadratic block predictors ($b = 14.25$, 95\% credible interval: $[11.40, 17.19]$; quadratic interaction: $b = -4.12$, 95\% credible interval: $[-6.64, -1.63]$). That is, participants selectively use fewer words to refer to tangrams about which they have communicated before, demonstrating that utterance shortening effects are not mere effects of fatigue. 

Predicting response times using the same model structure as above and a Gamma linking function appropriate for non-negative scales, we also found significant linear and quadratic block effects among the repeated condition tangrams ($b = -16.81$, 95\% credible interval: $[-18.52, -15.17]$; quadratic: $b = 2.60$, 95\% credible interval: $[1.27, 3.96]$) as well as interactions between both block predictors and repetition condition (linear block interaction: $b = 11.76$, 95\% credible interval: $[8.77, 14.69]$; quadratic block interaction: $b = -3.69$, 95\% credible interval: $[-6.60, -0.80]$) (\autoref{fig:stats}C). That is, participants became faster at choosing tangrams over the course of the experiment, and do so significantly more strongly for repeated tangrams.

So far, we have observed that dyads can communicate more efficiently and accurately throughout a game.
However, these coarse metrics are agnostic to the actual linguistic content of descriptions: the words used by the speaker to describe targets. 
To examine whether interlocutors are truly converging on shared conventions (despite swapping roles each block), we calculated the cosine similarity between OpenAI's text-embedding-ada-002 embeddings \cite{openaiEmbeddings} of the descriptions used on successive blocks for tangrams in the \emph{repeated} condition.

\begin{figure*}[!t]
\centering
\includegraphics[width=\textwidth]{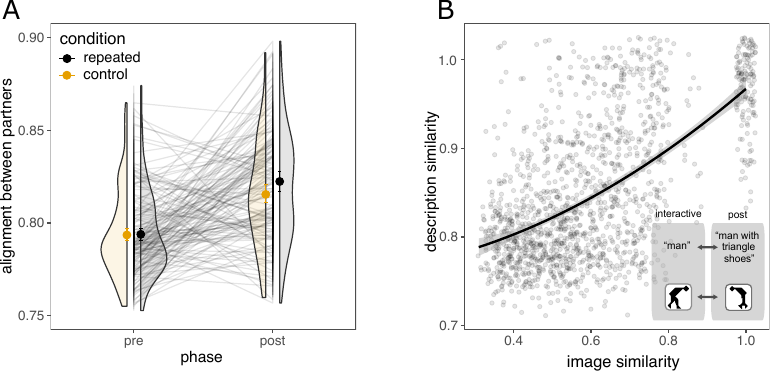}
\caption{(A) Cross-player description similarity in pre- and post-interactive phases, grouped by game. On average, partners show a larger increase in description alignment from pre- to post-interactive phase in the repeated condition (black) than control (yellow). (B) Same-player interactive vs. post-interactive phase description similarity as a function of image similarity, fitted with a quadratic function $y\sim poly(x,2)$. Tangram pairs are sampled from the same thread. The legend shows an example of one such pair. Description similarity increases nonlinearly with visual similarity. The error band indicates bootstrapped 95\% CI.}
\label{fig:fig-group}
\end{figure*}

We predicted cosine similarity between successive blocks' utterances with linear and quadratic effects of block and maximal by-dyad linear and quadratic random effects of block, using a Bayesian regression with a Gaussian link. We found a linear increase in similarity across successive blocks ($b = 0.47$, 95\% credible interval: $[0.41, 0.52]$) as well as a quadratic effect of block such that participants converged faster at first and leveled off later in the experiment ($b = -0.11$, 95\% credible interval: $[-0.16, -0.07]$). We also observed a similar trend for alignment of participants' exact lexical choices (see \aautoref{secA12}). 
These results indicate that participants converge on similar referring expressions with their partners, allowing shared conventions to stabilize (\autoref{fig:stats}D).

In sum, we showed that partners reliably developed referential conventions over the course of repeated interactions. They not only became more accurate and efficient in their communication, but also converged semantically on shared referring expressions. This result reproduced past observations in our pilot studies with similar setup (see \autoref{secA3} and \autoref{secA4}). This convergence suggests that dyads established stable conventions that could serve as a foundation for subsequent generalization beyond the repeated stimuli.

\subsection{Conventions generalize to new stimuli}

Do conventions formed while interacting generalize to items that were \emph{not} jointly discussed? To test this, we ask whether participants' descriptions of tangrams in the pre and post phases, which do not appear in the interactive phase, become more similar. Our prediction is that partners' descriptions of tangrams will become more similar after they interact, despite the fact participants never talked about these tangrams together.

To allow us to examine generalization, we designed our experiment with sets of three tangrams (tangram ``\textbf{threads}'') that were visually similar. For instance, all three tangrams in a thread may look like a person walking, but differ in other visual properties (see \autoref{fig:visualization}D).  
From each thread, two tangrams appeared in the pre and post phases and one appeared in the interactive phase. This design allows us to examine how participants' convergence on labels in the interactive phase leads to generalization to similar-looking tangrams that were not jointly discussed. Here, we examine participants' descriptions of the two undiscussed tangrams from each thread, modeling their change in similarity from the pre to the post phase. 

\begin{figure}[!t]
\centering
\includegraphics[width=\textwidth]{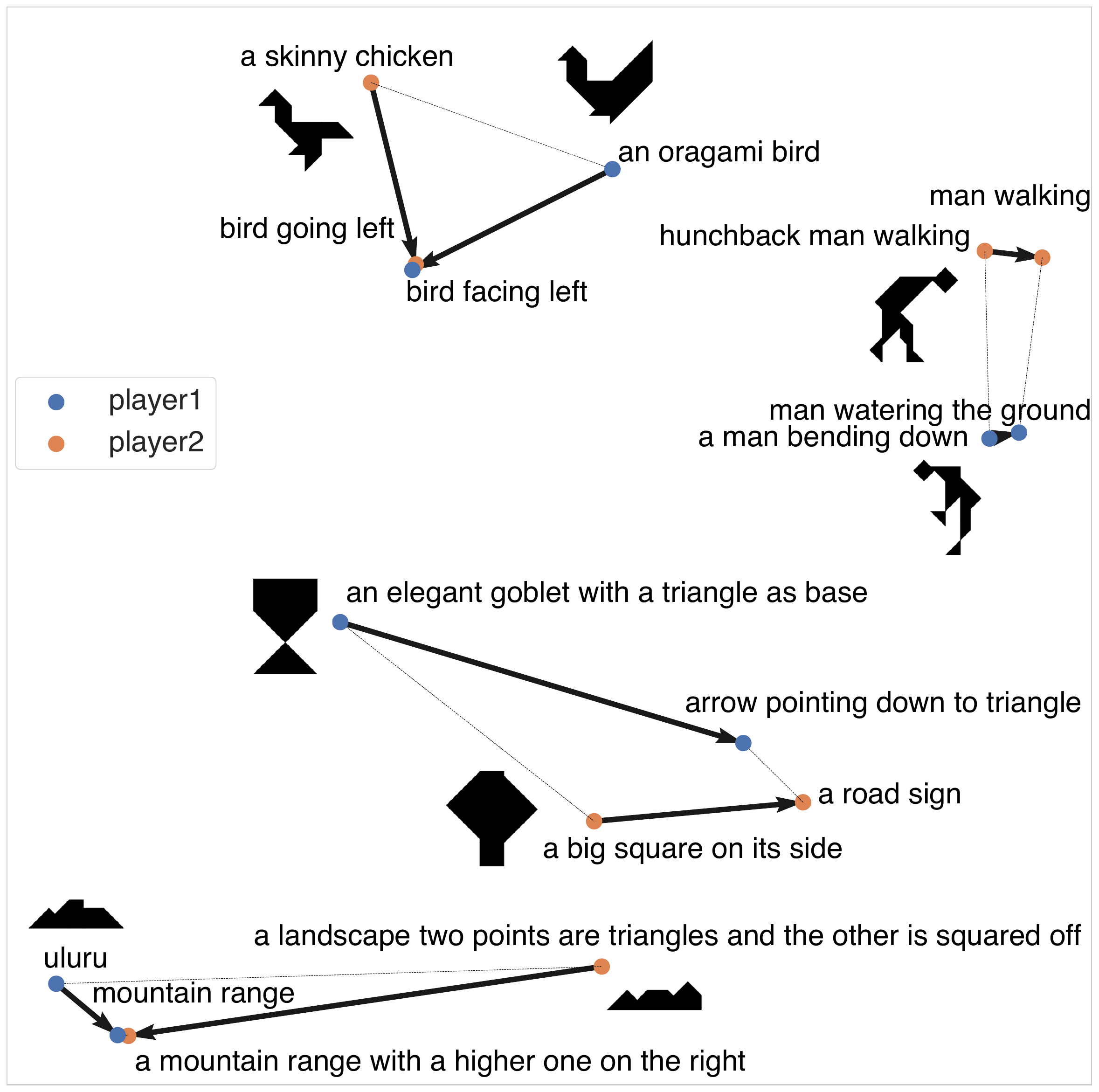}
\caption{Examples of cross-player description change from pre- to post-interactive phase in the semantic space. Text embeddings visualized using t-SNE \cite{JMLR:v9:vandermaaten08a} fitted on all the descriptions in all games. Within each pair, the two players each described a different tangram from the same (visually similar) thread.}
\label{fig:pre-post-diff}
\end{figure}

We modeled the similarity of dyads' descriptions of same-thread tangrams using phase, condition (repeated vs. control tangram), and their interactions as predictors, as well as maximal by-dyad random effects. 
The description similarity between same-thread tangram pairs increased from the pre to the post phase ($b=0.03$, 95\% credible interval: $[0.02,0.03]$). That is, players generalized their conventions from the interactive phase to new targets which were never jointly discussed  (\autoref{fig:fig-group}A). \autoref{fig:pre-post-diff} visualizes this alignment convergence between two players from pre to post phase in the semantic space.
For interaction between phase and condition, although the result is not significant ($b=0.01$, 95\% credible interval: $[-0.01, 0.00]$), the Bayes factor (BF$=15.53$) provides strong evidence \cite{lee2014bayesian} in favor of an interaction over the null model, with a posterior probability of $0.94$. This suggests a larger increase in description similarity for same-thread tangrams from threads repeated in the interactive phase. That is, players generalized conventions formed in repeated interactions to new targets more effectively than they did with non-repeated targets: repeated interactions established more generalizable conventions.

\subsection{Visual similarity mediates generalization}
According to the Universal Law of Generalization \citep{shepard1987toward}, in a psychological space established for a set of stimuli, the probability of generalizing a learned response or property of one stimulus to another decays exponentially with the distance between the pair. 
Our experiment showed that participants generalized the conventions formed in the repeated interactive phase to post phase tangrams. We want to investigate whether this generalization also follows the Universal Law, i.e., decays with the distance between the stimuli, as measured by visual similarity. If so, what is the form of this decay function?

To investigate participants' generalization in psychological space, we used participants' descriptions of pairs of tangrams from the same thread in the interactive phase and the post phase. We modeled participants' descriptive similarity from the interactive to the post phase as a function of the tangrams' visual similarity. Visual similarity was measured by cosine similarity between image embeddings from a CLIP model fine-tuned on black-and-white tangrams and annotations from \kilogram.

To examine the functional form of generalization with respect to visual similarity, we modeled this relationship with linear, quadratic, exponential, and combined quadratic and exponential functions. We conducted a model comparison to evaluate model fit by measuring the difference in expected log pointwise predictive density (ELPD) from the best fit model. We found that nonlinear models of generalization provide substantially better fit (\autoref{fig:fig-group}B), with the quadratic function providing the best fit and the combined quadratic and exponential model providing the second-best fit (ELPD difference $= -6.5$), followed by the exponential (ELPD difference $= -7.2$) and linear (ELPD difference $= -14.9$) models. We observed a similar trend when tangrams from different threads are included (see \autoref{si-fig: pre-post-sim}). 
There is a significant positive effect of both the quadratic ($b = 0.25$, 95\% credible interval: $[0.12, 0.38]$) and the linear ($b = 1.84$, 95\% credible interval: $[1.17, 1.96]$) term in the quadratic model, suggesting a convex upward relationship between label and image similarity. That is, we found that not only do participants converge on more similar labels for undiscussed tangrams in discrete ``threads'' of similarity, but they also do so in a graded manner, with more visually similar tangrams getting quadratically more similar descriptions. 
This aligns with recent findings \cite[e.g.,][]{marjieh2024universal}, in which quadratic models performed comparably to exponential ones in fitting the relationship between description and image similarities. In our case, the exponential fit is attenuated, likely due to greater referential ambiguity in abstract stimuli.

Additional analyses showed that description alignment between partners is mediated by the visual similarity to the interactive phase tangrams they formed conventions on (see \autoref{secA2}). After interaction, in the post phase, partners' description alignment is more strongly mediated by the visual similarity of new tangrams to those encountered during the interactive phase, indicating that interaction and convention formation amplify the role of visual similarity in generalization.

\subsection{Nameability facilitates communication}

\begin{figure}[t]
    \centering
    \fbox{\includegraphics[keepaspectratio=true, height=4.5em]{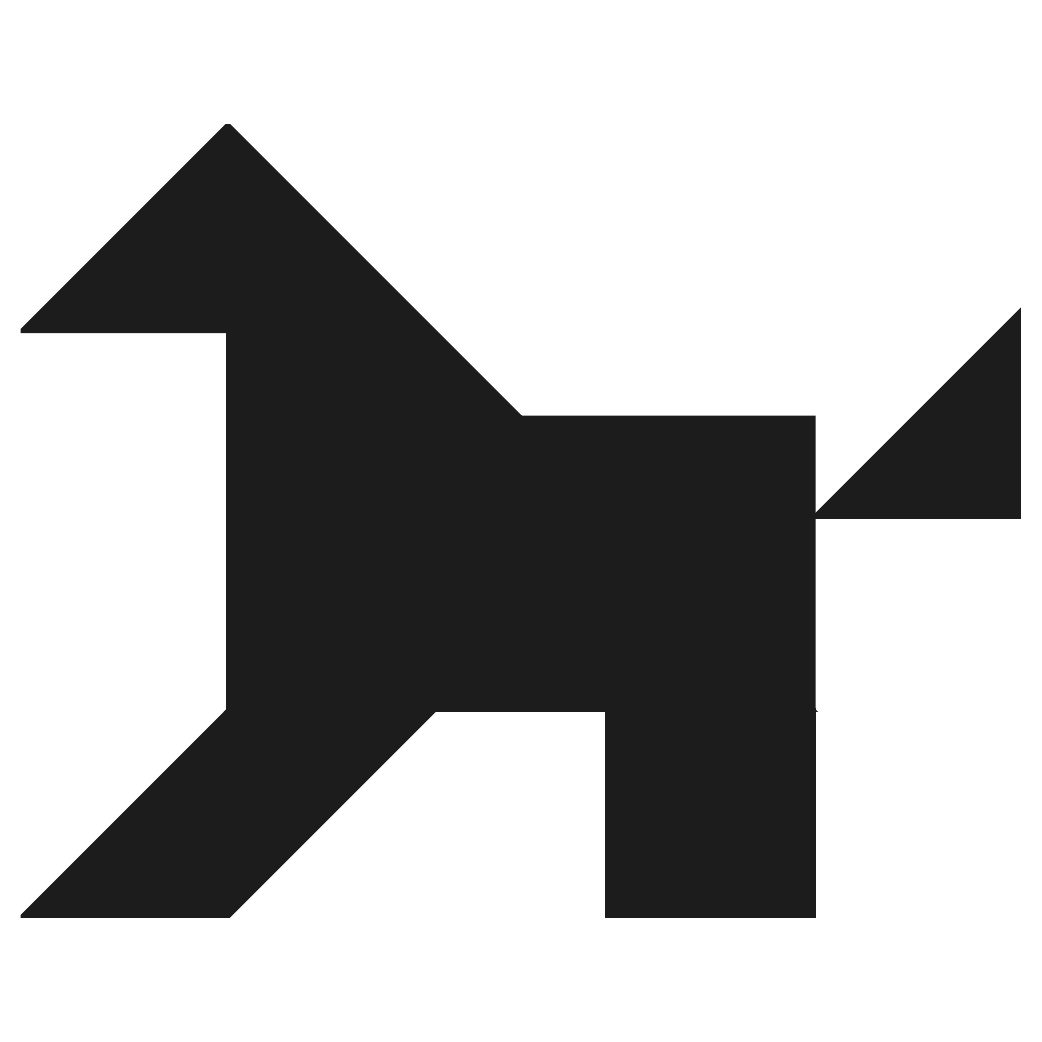}}\hspace{2em}\fbox{\includegraphics[keepaspectratio=true, height=4.5em]{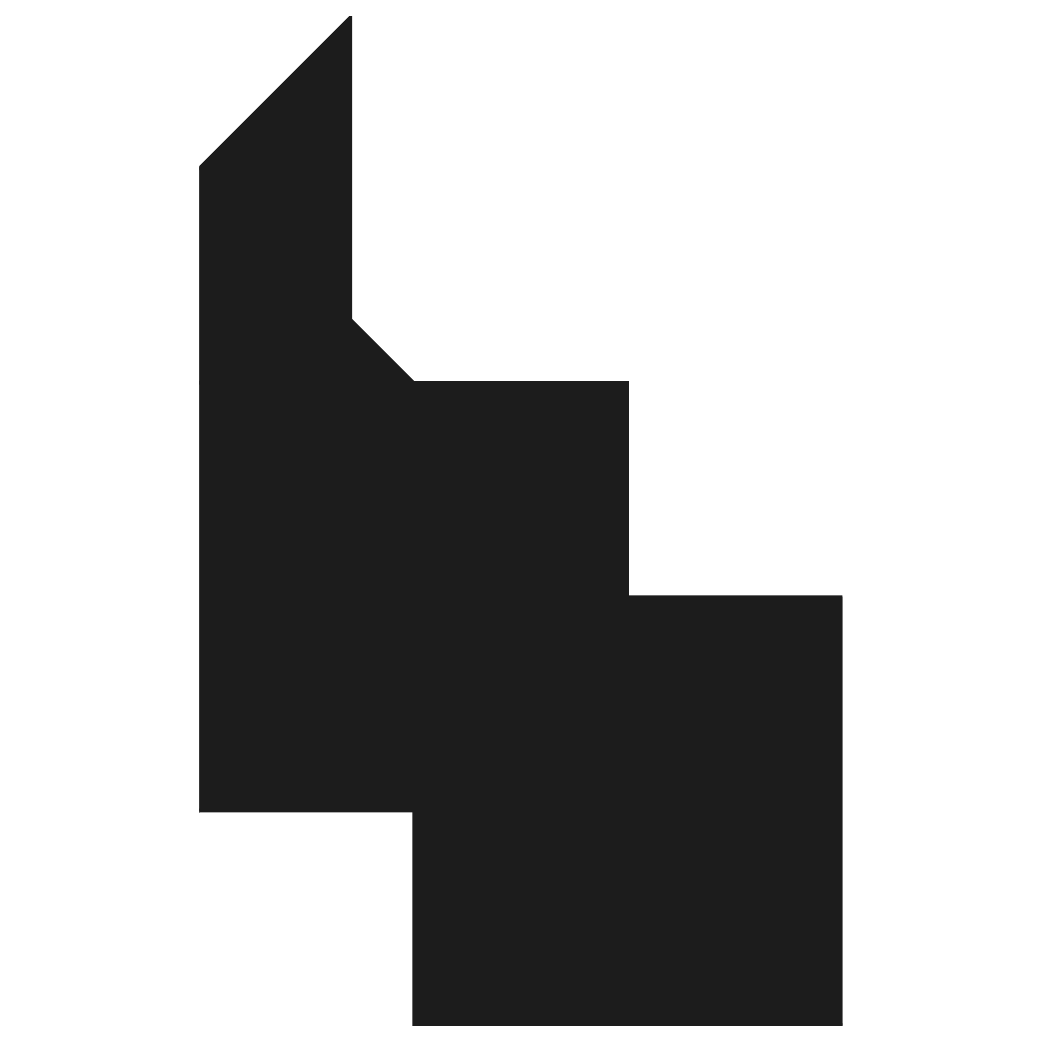}}
    \caption{Example tangram images from the \kilogram dataset. The one on the left has relatively high nameability, and the one on the right lower.}
    \label{fig:tangram-example}
\end{figure}

Dyads' interactions throughout the game clearly affect their descriptions, but their prior expectations about how to name things set the starting conditions for the words they choose. We investigate how people combine their prior expectations about naming with conventions formed in the task by considering the property of \emph{nameability} --- roughly, the \emph{a priori} likelihood that two individuals will prefer the same label before interacting~\citep{hupet1991effects, zettersten2020finding}.
For a higher-nameability referent, such as the left image in \autoref{fig:tangram-example}, most speakers will be expected to extend the same familiar label \emph{dog}.
Meanwhile, for a lower-nameability referent, such as the image on the right, different speakers offer very different labels (e.g., ``lizard'', ``robot hand'', or even ``lighter''). 
A recent study in the color domain has suggested that speakers make fairly accurate predictions about whether a given label will be shared by others \citep{murthy2022shades}, although this may not be the case for all domains \citep{lupyan2023hidden,marti2021latent,wang2021idiosyncratic}.

We measure nameability for each tangram using the additive inverse of Shape Naming Divergence (SND) \cite{Ji2022AbstractVR} (higher nameability is equivalent to lower naming divergence). We hypothesize that first, nameability may moderate the effects of interaction on naming similarity, and second, nameability will influence how efficiently dyads communicate, as reflected by accuracy, verbosity, and reaction time. We expect that high-nameability targets would be ``easier'' to communicate about than low-nameability targets across the board. 

We examine people's descriptions of tangrams in the pre and post phases to ask whether dyads use more similar descriptions for high-nameability tangrams, when those specific tangrams are not jointly discussed. Modeling the similarity of dyads' descriptions of same-thread tangrams using nameability, we found overall higher description similarity for more nameable targets ($b=0.13$, 95\% credible interval: $[0.10,0.16]$). We also found that change in description similarity did not interact with nameability---dyads generalized similarly across nameability levels, calculated as the average nameability of the three tangrams in the thread ($b=0.02$, 95\% credible interval: $[-0.04,0.07]$). That is, people start with and converge on more similar labels for high-nameability tangrams, but nameability does not significantly affect their \emph{change} in similarity.

We further evaluated effects of nameability throughout the interactive phase using a series of mixed-effects regression models for three aspects of communication performance: referential accuracy (whether or not the listener successfully selected the target), verbosity (the number of words in the speaker's descriptions), and speed (the time taken for the listener to make a selection after receiving the message).
For each metric, we construct a regression model including fixed effects of condition (repeated vs. control) and nameability class (high vs. medium vs. low), as well as an effect of block number (integers 1 to 5, centered) and all their interactions.
For accuracy (coded as a binary variable: correct vs. incorrect), we use a logistic linking function. 
Because all manipulations were within-dyad, we included the maximal random effect structure at the dyad level.

First, higher nameability facilitates accuracy. We observed a significant main effect of nameability on accuracy: all else equal, listeners were more likely to make errors for low-nameability targets than high-nameability ones ($b=2.33$, 95\% credible interval: $[0.45,4.52]$). Second, higher nameability allows dyads to communicate more succinctly: there is a significant main effect of nameability on description length ($b=-0.62$, 95\% credible interval: $[-0.89, -0.33]$), meaning higher-nameability targets received shorter descriptions.
Third, higher nameability allows dyads to identify referents faster: we found a significant main effect of nameability on listener response time, meaning listeners responded more quickly overall for high-nameability targets ($b=-0.46$, 95\% credible interval: $[-0.63, -0.28]$).
We also found a significant positive interaction between the linear component of block progression and nameability ($b=13.89$, 95\% credible interval: $[4.08, 24.04]$), indicating the effect of nameability on reducing response time weakens over blocks, especially for higher-nameability targets. This likely reflects a floor effect, since high-nameability targets require lower response time from the start of the interaction. In all, we found that higher nameability facilitates more accurate, succinct, and quick communication between dyads.

\section{Methods}\label{sec3}

\subsection{Participants}
We recruited 163 pairs of participants from Prolific based on preregistered inclusion criteria (English as first language and location based in US or UK). We excluded 12 pairs because their games were unfinished or contained more than 20\% empty responses, leaving 151 pairs ($N=302$) for analysis. Each game takes 45 minutes on average and the participants were awarded \$7 base pay and up to \$2.1 performance-based bonus.
 
\subsection{Stimuli}
We used black-and-white tangram shapes from the \kilogram dataset and constructed three contexts for each game. 
Each game in the experiment used 10 ``\textit{threads}'' of tangrams. A thread is a series of three tangrams, one for each context, i.e. 10 tangrams per context. We constructed a thread for each tangram (``head tangram'') with its two most similar tangrams in the dataset to connect the contexts based on visual similarity.
We ranked the threads by the similarity between the head tangram and its most similar tangram in the dataset and divided the range into 10 equal-sized bins, where the top two bins were combined because there were only four threads in the highest similarity bin. We sampled one thread from each of the nine bins, and the last thread contains three identical tangrams representing the similarity of 1. All tangrams within a game are unique.

We measured text and image similarity using cosine similarity (see \aautoref{secA11} for validation of these similarity metrics aligning with human similarity judgments). For texts, we calculated the cosine similarity between OpenAI's text-embedding-ada-002 embeddings \cite{openaiEmbeddings} of participants' descriptions. Compared to other commonly used text embeddings, text-embedding-ada-002 embeddings aligned best with human judgments of similarity (see \aautoref{secA11} for comparison details).
For images, we calculated the similarity between tangrams using cosine similarity between image embeddings from a CLIP model \citep{radford2021learning} fine-tuned on black-and-white \kilogram images and whole shape descriptions \citep{Ji2022AbstractVR}. The CLIP model encodes text and images separately and uses contrastive pre-training with symmetric cross-entropy loss. We obtained image embeddings using the image encoder of the fine-tuned CLIP model and calculated the cosine similarity between the embeddings.

\subsection{Design and procedure}
The experiment consisted of a consent form, a qualification quiz, three phases of reference games (pre, interactive, and post phases), a demographics survey, and a debrief form.
In each game, the interactive phase used a context (set of 10 tangrams) comprised of the head tangrams from 10 threads. Pre and post phases shared the same two contexts that each contained one tangram from the rest of each thread.

In the pre phase, the participants were asked to describe each of the 20 tangrams in the two contexts independently. They were shown one context at a time with the target highlighted and had 45 seconds to send one message to describe. The order of the two contexts and of tangrams in each context was random for the participants. After this phase, they were paired with a random partner to play a repeated reference game in the interactive phase. 

The interactive phase contained five blocks, and each block had six rounds. All rounds in this phase shared the same context. In this phase, five tangrams were assigned to the repeated condition and five to the control. Repeated tangrams appeared once as the target in every block while the controls appeared only once in one of the blocks as the target. The participants were randomly assigned to be the speaker or the listener and swapped roles after each block. The speaker needed to describe the highlighted target in the context to the speaker in 45 seconds, while the listener needed to select a tangram according to the description. The order of the tangrams in the context was randomized for the participants so that they could not rely on the location information in the description. Both participants were able to communicate freely through the chat box. If the timer had decreased to under 30 seconds, sending a message reset the timer back to 30 seconds so that both participants could communicate as much as necessary. After the listener selected a tangram, both participants would receive feedback on whether the target was correctly selected.

In the post phase, each tangram from the pre phase showed up as the target only once, and the participants took turns as the speaker to describe the target in 45 seconds. The listener was provided with an extra 15 seconds to make their selection if they hadn't already. The speaker could only send one message, and the listener could not reply. No feedback was provided after the selection. Each participant was guaranteed to describe one tangram in each thread.

\subsection{Data, materials, and software availability}
The experiment was approved by the Cornell University IRB. We implemented the experiment with Empirica \citep{Almaatouq2020EmpiricaAV}, a platform designed for creating and running real-time, interactive online experiments involving human participants. The experiment was hosted on an Amazon EC2 instance, and all the data was stored in MongoDB. All code and data are publicly available at: 
\hyperlink{https://github.com/lil-lab/tangrams-ref}{https://github.com/lil-lab/tangrams-ref}.

\section{Discussion}\label{sec4}
When we form conventions with others, what exactly do we form? Using tangrams from the \kilogram dataset, a large-scale high-diversity resource for abstract stimuli, we designed a three-phase reference game experiment to track convention formation through repeated interactions and investigate how people generalize these conventions to new entities. 
We observed that conventions, manifested through stable shared descriptions and more efficient communication, are not confined to the entities they were coined to name. Partners showed increased alignment when describing undiscussed entities, suggesting genuine conceptual coordination rather than item-specific memorization. 
Moreover, the strength of participants' generalization was modulated by visual similarity: consistent with the Universal Law of Generalization \citep{shepard1987toward}, the strength of generalization scales nonlinearly with visual similarity.
While nameability affected baseline description similarity, with high-nameability objects receiving more similar descriptions throughout, it did not modulate the degree of alignment, indicating that conceptual coordination operates similarly across different levels of prior consensus.

These findings have implications for theories of convention formation and the cognitive mechanisms driving the emergence and evolution of systems of reference. 
While recent work has shown that conventions formed with one partner gradually generalize to new partners through hierarchical inference \cite{hawkins2022partners}, our results reveal a complementary dimension of generalization to undiscussed but visually similar targets. 
This axis of generalization may illuminate how linguistic categories expand over time \cite{xu2016historical,ramiro2018algorithms} as meanings are ``chained'' from one referent to related ones. 
This process also appears fundamental to language acquisition: children systematically overextend words like ``dog'' to novel referents based on similarity \cite{clark1973s}, and recent models suggest that word meanings may be dynamically constructed through efficiency-driven chaining \cite{yu2025infinite}. 
Thus, the ad hoc conventions formed in momentary social coordination may serve as a microcosm for understanding both developmental and cultural-evolutionary processes that shape linguistic systems.

These insights into human conceptual alignment present both challenges and opportunities for building adaptive artificial systems that can interact effectively with human partners. 
Current approaches to grounded language learning have made progress on specific aspects of convention formation: extracting reference chains from dialogue history \citep{haber2019photobook, takmaz2020refer} and establishing and maintaining a common ground in a specific context \citep{chai2014collaborative, udagawa2021maintaining}. 
However, these systems typically treat conventions as fixed mappings between expressions and referents, missing the levels of ad hoc alignment that humans are able to achieve across both partners and targets.
This suggests that human-like language agents may require architectures that go beyond episodic mappings to instead update their representations to dynamically align with partners throughout interaction \citep{suhr2022continual, hawkins2019continual}, consistent with recent proposals that synthesize situation-specific representations on-the-fly \cite{ying2025language,wong2025modeling}. 

Our study raises several important directions for future work. First, contextual factors modulate this process. For example, high-nameability contexts might enable rapid but shallow coordination, while low-nameability environments could drive deeper conceptual alignment. The set of alternative referents may also influence convention formation and transfer: distinctions that matter in one context may become irrelevant in another, affecting how conventions generalize. 
Second, while we studied abstract referents as a tractable window into these processes, the mechanisms likely extend beyond object reference to coordination on beliefs, attitudes, and shared goals: wherever successful interaction requires conceptual alignment. Future work should focus on real-world settings, from classrooms where teachers and students must establish shared understanding, to clinical contexts requiring patient-provider alignment, to the challenge of building AI systems that can dynamically adapt their semantic representations.
Ultimately, the conventions that undergird our language---the meanings of our words and the procedures by which we transform them into meaningful sentences---all started as ad hoc conventions. These findings lay the groundwork to explain how our momentary attempts to align with interlocutors become fully-fledged language systems over language development and language evolution.

\backmatter


\bmhead{Acknowledgements}

This research was supported by NSF under grant No. 1750499 to YA and No. 2404676 to CAB, and a gift from Open Philanthropy to YA. RDH was supported by a C.V. Starr Fellowship. We are grateful for helpful conversations with Boris Harizanov and for the contribution from Prolific workers.





\bibliography{sn-article}

\begin{appendices}

\section{Validating similarity metrics}\label{secA1}
\subsection{Embedding similarity alignment with human judgment}\label{secA11}
We ran an additional experiment to collect human voting on similarity to ensure that the metrics we used to measure text and image similarities align with human perception. We used CLIP \citep{radford2021learning} fine-tuned on \kilogram \citep{Ji2022AbstractVR} as the embedding model for image pair similarity and experimented with 3 different text embeddings for text pair similarity: Sentence-BERT (SBERT) \citep{reimers-2019-sentence-bert}, fine-tuned CLIP, and OpenAI's text-embedding-ada-002 (Ada) \citep{openaiEmbeddings}. 
We measured the similarity of pairs using cosine similarity between the embeddings and compared the results with human judgments by voting for the more similar pair between two pairs. 

\subsubsection{Participants} We recruited 63 native English speakers based in the US or UK from Prolific (31 for text similarity and 32 for image similarity). Each participant was tasked to compare 50 pairs of similarities, which took 9 minutes on average. They were awarded \$2 for completing the task.

\subsubsection{Stimuli} To validate text pair similarities, we used 1472 tuples of utterances from the experiment. Each tuple consisted of two pairs of descriptions from the same game, one from the pre-interactive phase and the other from the post-interactive phase. Each pair consists of player 1's description of tangram $A_i$ and player 2's description of tangram $B_i$, where tangram $A_i$ is from set $A$ in the same thread $i$ as tangram $B_i$ from set $B$. 
We used 1474 tuples of two image pairs to validate image pair similarities. Each image pair used one tangram from the interactive phase and one from the post-interactive phase, either from the same thread or randomly sampled from a different thread from the post-interactive phase. 

\begin{figure}[t]
\centering
\includegraphics[width=\textwidth]{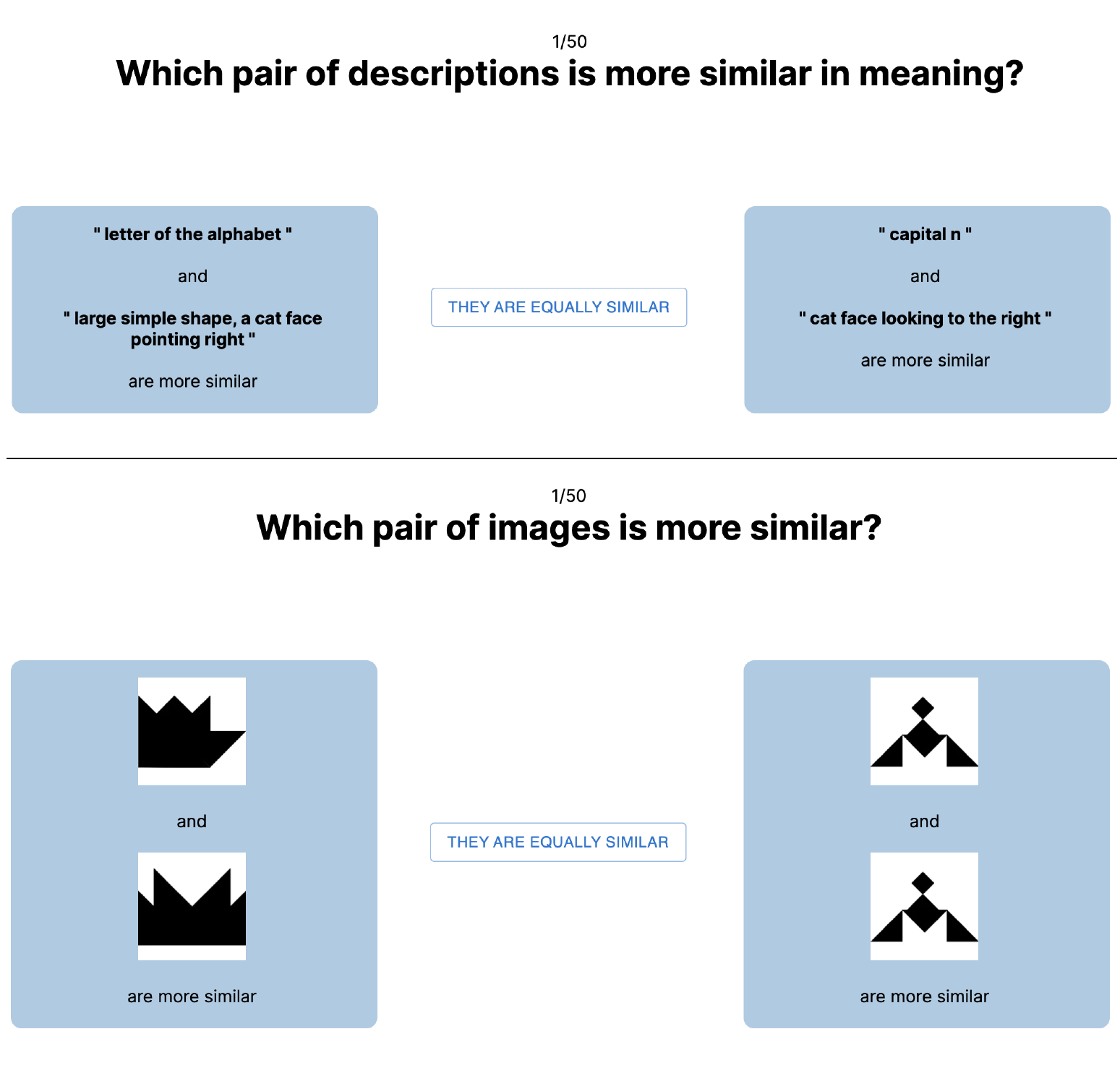}
\caption{Annotation interface for collecting human votes on pair similarity between texts (top) and images (bottom).}
\label{si-fig: validate}
\end{figure}

\subsubsection{Method}
The participants were asked to read the instructions and complete three practice trials correctly to move on to the actual trials. \autoref{si-fig: validate} shows the interface for the actual trials. The participants were prompted to click on the more similar pair or select ``they are equally similar''. We calculated the cosine similarities with the embeddings of the two pairs, computed the difference by subtracting the second pair's similarity from the first pair's similarity, and compared them with the responses from the human participants to see if they were consistent.

\subsubsection{Result}
We collected a total of 1587 responses for 1474 text pairs and 1593 responses for 1494 image pairs from human participants, with at least one vote for each pair.
We took the majority vote as the human judgment if there was more than one vote for a pair and excluded the pair if there was a tie, resulting in 1447 text pairs and 1458 image pairs for analysis. A human response of ``equally similar'' was considered correct when the cosine similarity difference between the two pairs was less than 0.1. Ada embeddings showed the highest alignment with human judgments (83.6\%), compared to SBERT (66.1\%) and CLIP (64.8\%). Fine-tuned CLIP image embeddings also showed relatively strong alignment with human judgment, achieving 71.1\% accuracy.

Moreover, we used Bayesian ordinal regression with a cumulative logit link to model human judgments as a function of cosine similarity differences from each embedding model. Separate models were fit for each embedding type, and model performance was compared using approximate leave-one-out cross-validation, based on differences in expected log predictive density (ELPD). 
For text embeddings, the model using Ada similarity differences provided the best fit to human judgments, outperforming SBERT (ELPD difference = –5.1) and CLIP (ELPD difference = –15.6). The posterior for Ada's effect estimate was strongly positive ($b = 18.70$, 95\% credible interval: $[16.57, 21.04]$), indicating that increases in Ada embedding similarity aligned reliably with higher similarity by human judgment. 
For image embeddings, the model using fine-tuned CLIP image embeddings revealed a strong positive effect of cosine similarity difference on human judgment ($b = 4.11$, 95\% credible interval: $[3.77, 4.45]$), indicating that increases in image similarity were reliably associated with upward shifts in human judgments (from ``less similar'' to ``more similar'').

\begin{figure}[t]
\centering
\includegraphics[width=\textwidth]{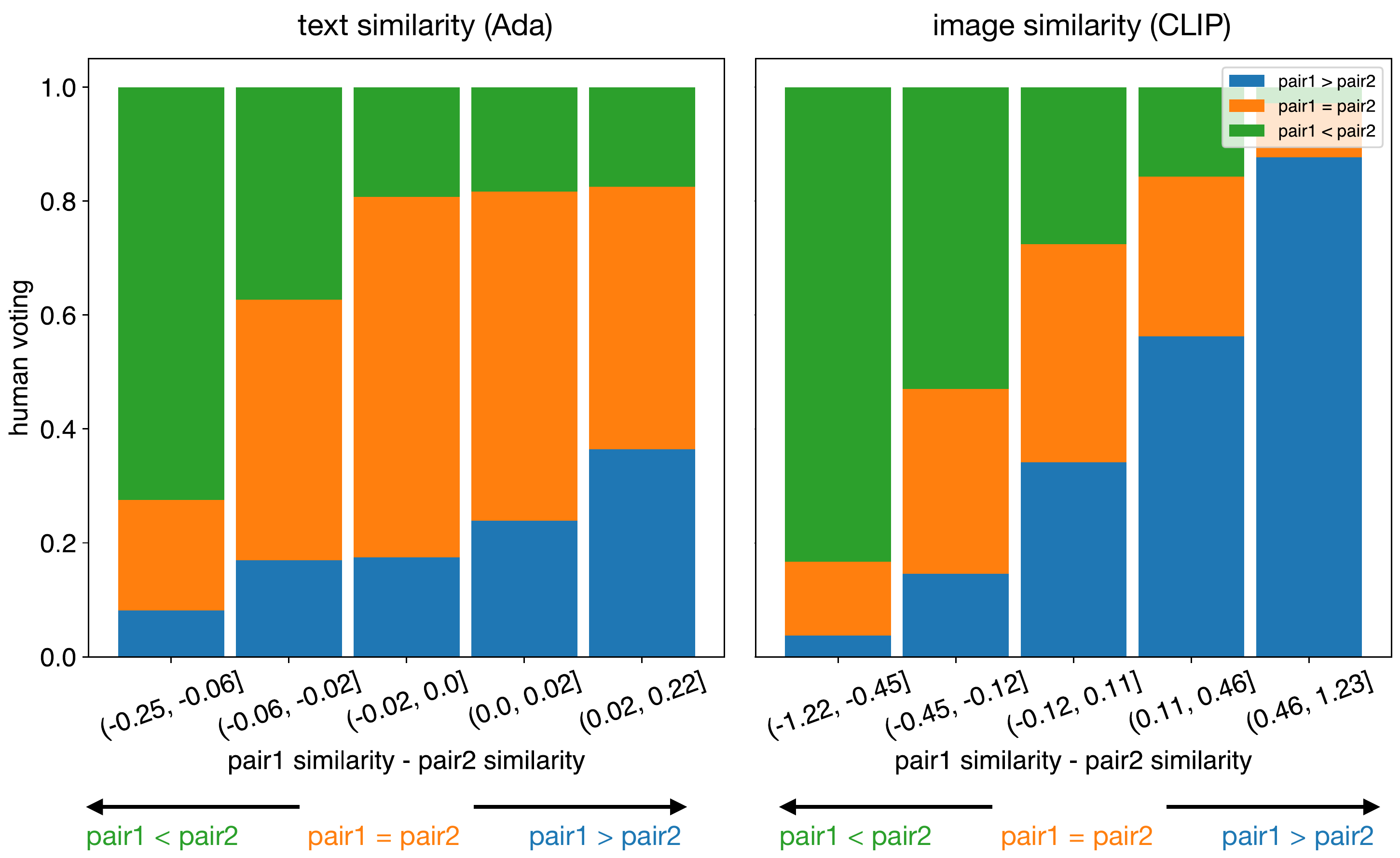}
\caption{Cosine similarity calculated with embeddings compared with human voting of similarity between text description pairs (left) and image pairs (right). Human voting result is shown in percentages. Each bin contains the same number of responses. \texttt{pair1 > pair2} means \texttt{pair1} of texts or images is more similar than \texttt{pair2}, and vice versa.}
\label{si-fig: human-compare}
\end{figure}

These results confirmed that Ada text embeddings and fine-tuned CLIP image embeddings showed strong alignment with human judgment of similarity, supporting their validity as measures for quantifying perceived similarity in the experiment.
\autoref{si-fig: human-compare} shows a qualitative comparison between human voting and the difference in the cosine similarity between text or image pairs, as calculated using Ada embeddings for texts and CLIP embeddings for images. 

\subsection{Jaccard similarity for lexical alignment between participants}\label{secA12}
Cosine similarity between embeddings measures the semantic alignment between texts, while Jaccard similarity provides a stricter metric based on exact lexical overlap. The Jaccard index is a measure of set similarity defined as the size of the intersection over the size of the union, e.g. $$J(W_1, W_2) = |W_1 \cap W_2| / |W_1  \cup W_2|$$
To ensure effects were not driven by spurious changes in function words or pluralization, we lemmatized all speaker descriptions and excluded stop words before computing sets.

\begin{figure}[t]
\centering
\includegraphics[width=\textwidth]{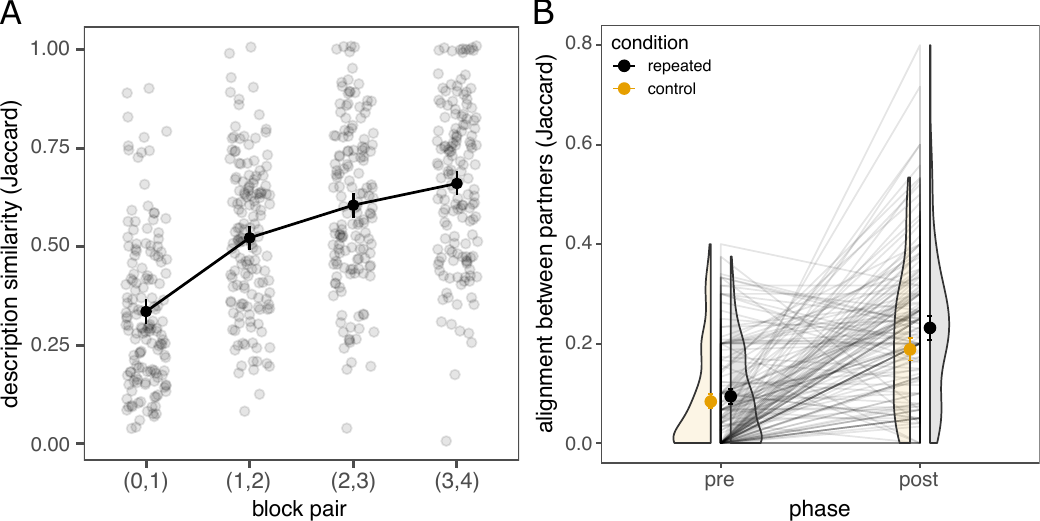}
\caption{(A) Participants increasingly aligned on the same descriptions, where the similarity between the descriptions used on successive blocks is measured by Jaccard similarity. Error bars are bootstrapped 95\% CIs and curves are best-fitting quadratic regression fits. (B) Cross-context cross-player pre- vs. post-interactive phase description similarity difference, grouped by game. The orange line shows the average increase in similarity from pre- to post-interactive phase.}
\label{si-fig: jaccard}
\end{figure}

We measured lexical alignment for both the interactive phase and between pre- and post-interactive phases. 
During the interactive phase, participants increasingly aligned on the same descriptions both in semantics and on the lexical level. We measured the similarity between the descriptions used on successive blocks using the Jaccard index, $J(W_i, W_{i+1}$), between the set of words $W_i$ used on successive blocks for tangrams in the \emph{repeated} condition.
We found a steady increase in similarity across successive blocks (\autoref{si-fig: jaccard}A), indicating that participants increasingly reused exact lexical choices from their partner in the previous block as shared conventions stabilized.\footnote{We found a similar effect for a directed variant of the Jaccard index: $J_{dir}(W_1, W_2)=|W_1 \cap W_2| / |W_2|$ normalizing by the size of the later description rather than the union. This supplemental analysis suggests that the observed changes in set similarity were not driven by non-stationarity in set size over time.}

We predicted Jaccard similarity between successive blocks' utterances with linear and quadratic effects of block and maximal linear and quadratic block by-dyad random effects, using a Bayesian regression with a Gaussian link. We found a linear increase in similarity across successive blocks ($b = 2.89$, 95\% credible interval: $[2.60, 3.18]$) as well as a quadratic effect of block such that participants converged faster at first and leveled off later in the experiment ($b = -0.80$, 95\% credible interval: $[-1.05, -0.56]$). These results indicate that participants converge on similar referring expressions with their partners, allowing shared conventions to stabilize.

In addition, from pre- to post-interactive phase, participants showed higher Jaccard similarity to their partner describing the same tangrams after forming conventions on a different set in the interactive phase (\autoref{si-fig: jaccard}B). This suggests that the players were able to generalize the convention beyond context by using similar word choices on new targets in the post-interactive phase.

\section{Interaction amplifies visual similarity effect on generalization}\label{secA2}
\begin{figure}[t]
\centering
\includegraphics[width=\textwidth]{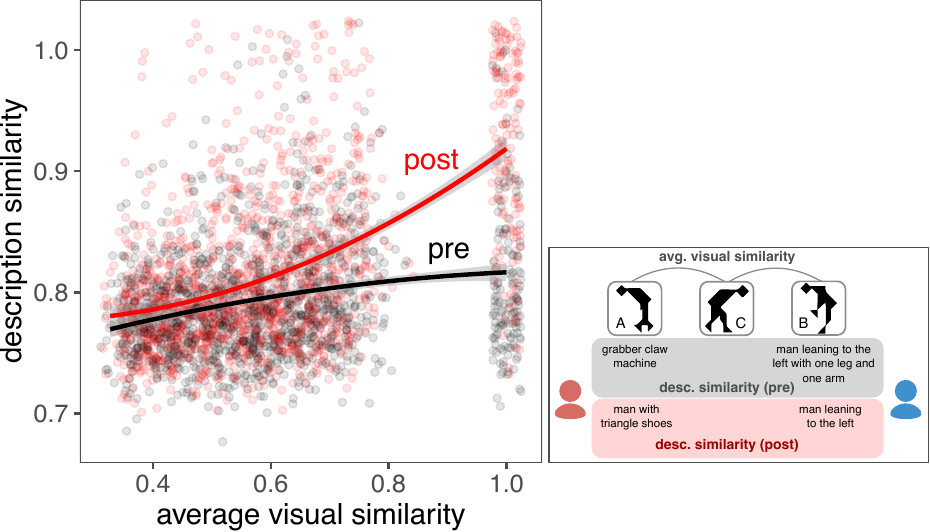}
\caption{Cross-player pre- (black) and post-interactive (red) phase description vs. average visual similarity to the interactive phase tangram, fitted with a quadratic function $y\sim poly(x,2)$. The error band indicates bootstrapped 95\% CI. After interaction, in post phase, the description similarity between two players for same-thread tangrams is more strongly mediated by the tangrams' average visual similarity to the interactive phase tangram from that thread.}
\label{si-fig: pre-post-sim}
\end{figure}
In the main text, we found that after forming conventions in the interactive phase, partners' description alignment increases in the post phase compared to the pre phase. Additionally, we found that individual player's description similarity increases non-linearly with visual similarity. In this section, we also show that label alignment is mediated by the visual similarity to the interactive phase where conventions were formed, and that the interaction amplifies this effect on label alignment between two players.

To test how visual similarity to the interactive phase stimuli might affect label similarity between players, we modeled each pair of players' description similarity as a function of the average visual similarity of pre/post phase tangrams to the interactive phase tangrams in the same thread, split by pre and post phase. 
We modeled this relationship with linear, quadratic, and exponential models and conducted model comparison using approximate leave-one-out cross-validation. 
In the pre phase, the quadratic model provides the best fit, significantly outperforming the next-best linear model (ELPD difference = $-26.3$. The quadratic model has a significant positive linear term ($b = 0.46$, 95\% credible interval: $[0.36, 0.57]$) and a modest negative quadratic term ($b =-0.10$, 95\% credible interval: $[-0.21, -0.00]$), suggesting a weak negative effect of the average visual similarity on label alignment. 
In the post phase, the quadratic model also provides the best fit with the exponential model as the second best fit (ELPD difference = $-11.3$). There is a significant positive effect of both the linear ($b = 1.42$, 95\% credible interval: $[1.28, 1.56]$) and quadratic term ($b = 0.32$, 95\% credible interval: $[0.19, 0.45]$) in the best fit model, indicating a stronger and accelerating effect of visual similarity on label alignment after the interactive phase (\autoref{si-fig: pre-post-sim}). 

To show that the interaction amplifies the influence of visual similarity on how players generalize and coordinate linguistic descriptions, we fit a mixed-effects model predicting label similarity from average visual similarity, phase (pre/post), and their interaction. 
The main effect of visual similarity was positive and significant ($b = 0.21$, 95\% credible interval: $[0.19, 0.23]$), indicating that on average, higher visual similarity to the interactive phase tangram was associated with greater alignment in player descriptions. There is also a significant main effect of phase ($b = -0.03$, 95\% credible interval: $[-0.03, -0.02]$), suggesting that label similarity was slightly lower in the pre phase compared to the post phase overall.
There is a significant negative interaction effect between phase and visual similarity to the interactive phase ($b = -0.14$, 95\% credible interval: $[-0.16, -0.12]$), indicating that the effect of visual similarity on label alignment was weaker in the pre phase. These results confirm that after establishing conventions through interaction, players become more attuned to visual similarity when generalizing their linguistic labels to new tangrams.

\begin{figure}[t]
\centering
\includegraphics[width=\textwidth]{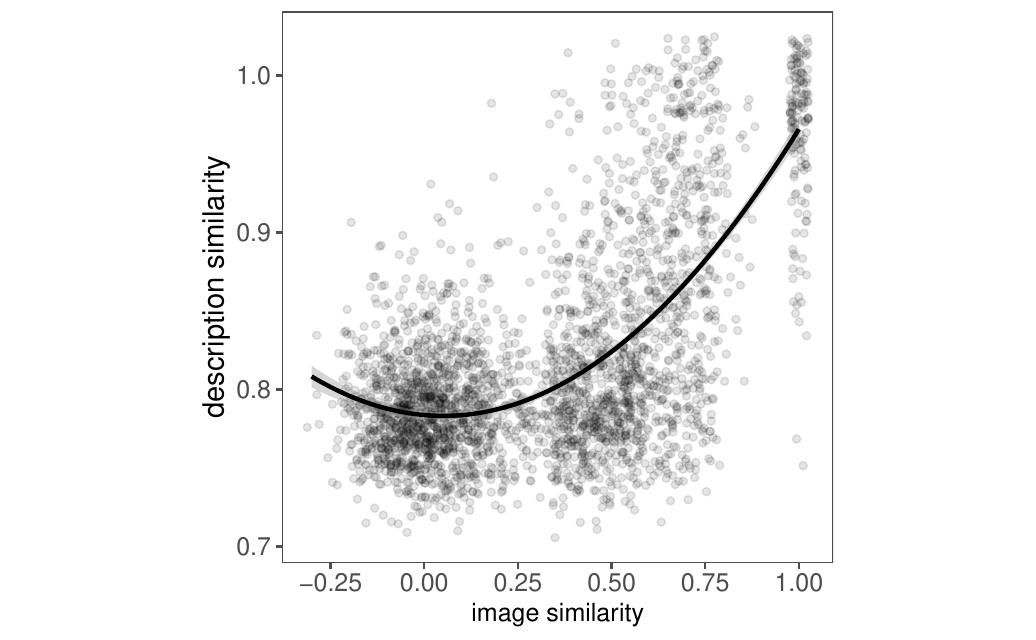}
\caption{Same-player interactive vs. post-interactive phase description similarity as a function of image similarity, fitted with a quadratic function $y\sim poly(x,2)$. Tangram pairs are sampled from the same thread and across different threads. Description similarity increases nonlinearly with visual similarity. The error band indicates bootstrapped 95\% CI. Quadratic function provides the best fit, and the combined quadratic and exponential model provides the second-best fit (ELPD difference $= -7.0$), substantially preferred over exponential (ELPD difference $= -123.0$) or linear (ELPD difference $= -259.7$) models.}
\label{si-fig: shepard_diff_thread}
\end{figure}

\section{Pilot 1: Manipulating nameability}\label{secA3}

\subsection{Participants}
We recruited 60 pairs of participants from Prolific, based on preregistered inclusion criteria (English as first language and location based in US or UK). We excluded 8 pairs of participants, because their games contained more than 20\% empty responses. Participants provided informed consent in accordance with the institutional IRB. Each game lasted an average of 23 minutes and participants were given a base payment of \$4.25 (approximately \$11 per hour) with a performance bonus up to \$0.90.

\subsection{Stimuli}
We designed a reference game using the black-and-white tangram shapes from the \kilogram dataset. 
Each shape in \kilogram was previously normed for nameability using a metric called Shape Naming Divergence (SND), computed over naming annotations included in the dataset. 
This metric is defined as the mean proportion of words in each description that do not appear in any other description for that tangram. 
For example, if all annotators of a tangram used the one-word description ``bird'', the SND would be 0 because there are no unique words; if all annotators used distinct one-word descriptions, the SND would be 1. 
We sorted all 1016 tangrams in the dataset by SND and selected the top 100 tangrams to use for the \emph{low-nameability} condition and the bottom 100 for the \emph{high-nameability} condition to ensure maximal differentiation.

For each pair of participants, we sampled a context of 10 tangrams, 5 tangrams from the high-nameability set and 5 tangrams from the low-nameability set.
We ensured that tangrams in the context are sufficiently distinct to avoid challenging informativity pressures (e.g. two high-nameability tangrams that both have the consensus label \emph{bird}).
For each tangram, we extracted the head word from each of the 10 \textsc{KiloGram} descriptions using SpaCy v3 \cite{spacy} dependency parser.
We then set a threshold of $10\%$ overlap in any pairwise list of head words.
We reject and re-sample sets with pairs above the threshold. 

\subsection{Design and procedure}
\begin{figure*}
\centering
\includegraphics[width=\textwidth]{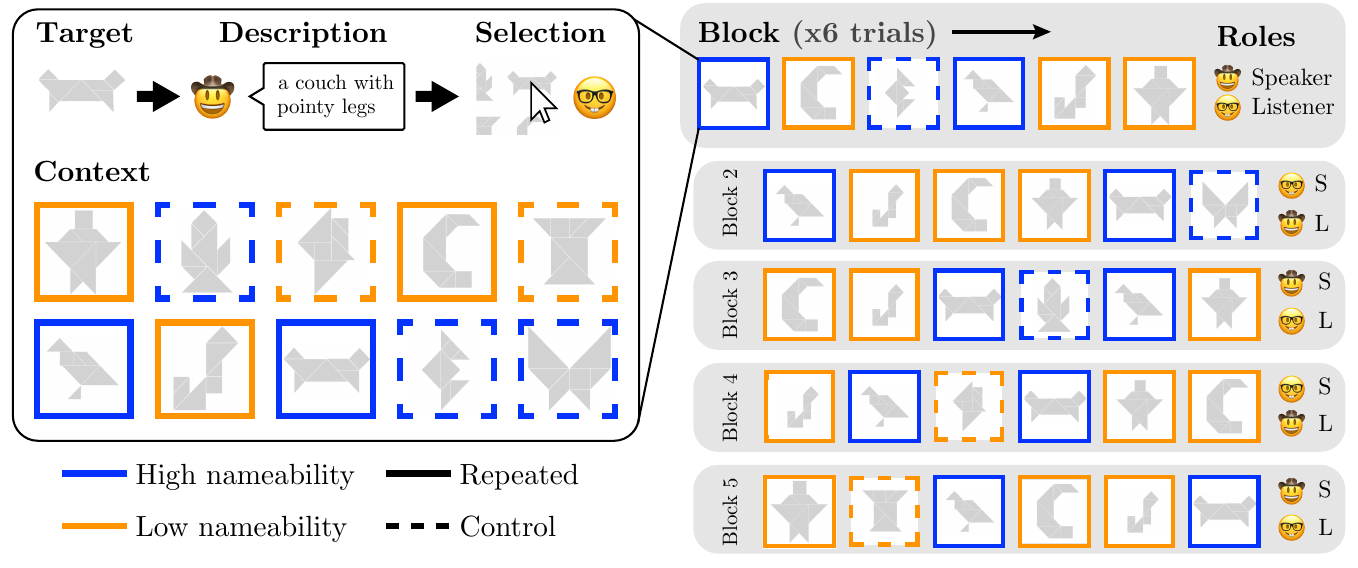}
\caption{\textbf{Pilot 1 design}. On the left, we show an example of a single reference game trial, using a mixed context of ten tangrams (borders added for this graphic). The context remains the same for all blocks and trials. On the right, we show a full target sequence consisting of five blocks. In each block, we randomly inserted a single control tangram target (dashed border) among the repeated targets (solid border).}
\label{si-fig: exp1}
\end{figure*}
\autoref{si-fig: exp1} illustrates the design. Each game contained 5 blocks, and each block had 6 trials. All trials in a game were based on the same context. 
In each context, 5 tangrams were assigned to the \emph{repeated} condition, each appearing exactly once in each block. Among the 5 repeated tangrams, we ensured that 2 were drawn from the low-nameability condition and 3 from the high-nameability condition, or vice versa. The other 5 tangrams were assigned to the \emph{control} condition and only appear in one of the blocks as the target. 
Thus, all experimental manipulations were within-dyad. 
The order of targets are randomized in each block, and participants alternated roles between blocks. 

Participants were randomly assigned to pairs after providing consent and passing a tutorial and a quiz. 
The participants were randomly assigned to the speaker or the listener roles.
When the game started, both players saw 10 tangrams and a chat box. 
The order of the tangrams was different from the speaker's and the listener's views, so that the speaker could not rely on the position when describing the target. The speaker was asked to send only one message to describe the highlighted target tangram from the context in 45 seconds. The listener needed to select a tangram based on the description. An additional 15 seconds were given to the listener to make the selection if they had not already. Both participants received feedback after each trial indicating if the listener had responded correctly. 
At the end of the experiment, the participants were given a demographic survey about their age, gender, language background, game experience, feedback for the study, etc.
The experiment was built with Empirica, a platform for building and conducting synchronous and interactive online experiments with human participants. 
We hosted the games on Meteor Cloud and stored game data in MongoDB.

\subsection{Results}

We tested three key hypotheses about performance over time. 
First, we expected that high-nameability targets would be ``easier'' to communicate about than low-nameability targets across the board.
Second, we expected that performance would improve overall for repeated targets to a greater degree than the non-repeated controls interspersed throughout the trial sequence.
Third, and of greatest theoretical interest, we ask whether there is a three-way interaction: the performance improvements that accrue over successive interactions for repeated targets may more readily transfer to improved performance on low-nameability controls than high-nameability controls.

We evaluated these predictions using a series of mixed-effects regression models for three complementary metrics of communication performance: referential accuracy (whether or not the listener successfully selected the target), verbosity (the number of words in the speaker's description), and speed (the time taken for the listener to make a selection after receiving the message).
For each metric, we construct a regression model including fixed effects of condition (repeated vs. control) and nameability class (high vs. low), as well as an effect of block number (integers 1 to 5, centered) and all their interactions.
For accuracy (coded as a binary variable: correct vs. incorrect), we use a logistic linking function. 
Because all manipulations were within-dyad, we included the maximal random effect structure at the dyad-level.

\begin{figure}[t]
\centering
\includegraphics[width=\textwidth]{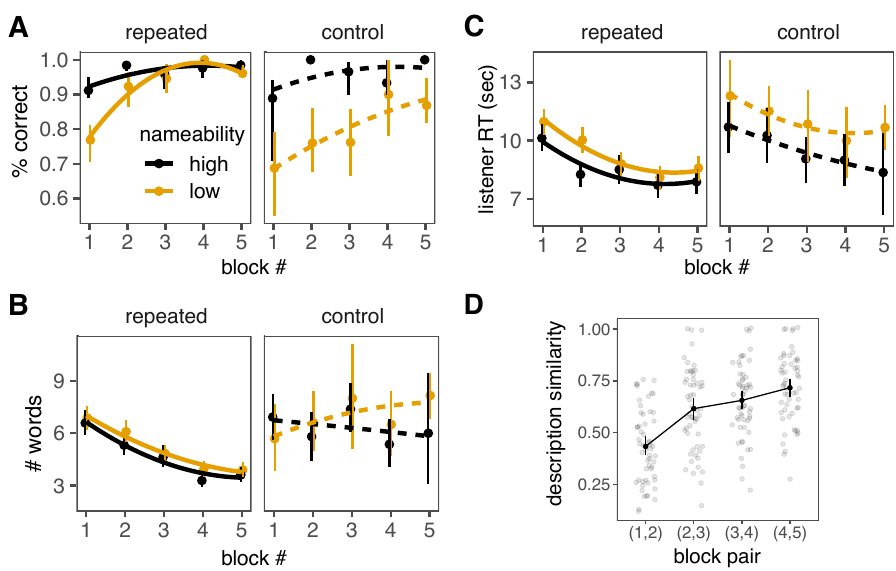}
\caption{\textbf{Pilot 1 results}. (A) accuracy, (B) description length, and (C) time elapsed between speaker message and listener response (in seconds) for low- and high-nameability tangrams in the repeated (left) or control (right) condition. Error bars are bootstrapped 95\% CIs. (D) To evaluate the stability of referring expression content over the course of the game, we computed the Jaccard index between the set of all words used in each pair of successive blocks. We found a gradual increase in the metric, indicating increasing block-to-block similarity in word sets.}
\label{si-fig: exp1_results}
\end{figure}

\subsubsection{Accuracy}
Our raw accuracy metric is shown in \autoref{si-fig: exp1_results}A.
First, we observed a significant main effect of nameability: all else equal, listeners were more likely to make errors for low-nameability targets than high-nameability ones, $b=0.87, z=-3.4, p < 0.001$.
While success rates improved overall over the course of the task, $b = 33.3, z=4.1, p < 0.001$, the most complex model supported by our data only included a significant interaction between condition and nameability, $b=-0.32, z=-2.06, p = 0.039$, likely reflecting ceiling effects for high-nameability targets, and no significant interaction was found with block number.

\subsubsection{Description length}
We considered the efficiency with which speakers were able to communicate, measured as the number of words in their descriptions (\autoref{si-fig: exp1_results}B).
We first found an overall main effect of block, $b=-18.6,~t(105) = -3.7,~p<0.001$, consistent with classic observations~\citep{krauss1964changes,clark1986referring}, as well as a significant main effect of nameability, $b=0.38,~t(47)= 2.8,~p=0.008$, where high-nameability targets received shorter descriptions.
We also found a significant interaction between condition and block: holding nameability constant, the decrease in utterance length for repeated tangrams was significantly greater than for control tangrams, $b = 25.2,~t(689)=6.5,~p<0.001$.
Finally, we found a significant three-way interaction, $b = 8.4,~t(1066)=2.1,$ $p=0.039$, consistent with the hypothesis that dyads are able to converge on shorter descriptions for repeated tangrams regardless of their nameability, but these improvements only generalize to high-nameability control tangrams.
If anything, when speakers must refer to a low-nameability control tangram for the first time late in the game, they produce slightly longer expressions than they did earlier in the game.

\subsubsection{Listener response time}
Our third performance metric is the time required for listeners to respond after receiving a description (\autoref{si-fig: exp1_results}C), as listeners may be expected to pause longer when uncertain about their response.
To ensure that listener response times are not driven by the time window left by speakers, we capped response times at the maximum value of 15 seconds given to listeners on trials when the speaker ran out the clock.
We found a significant main effect of nameability, where listeners responded more quickly overall for high-nameability targets, $b=0.63,~t(139) = 4.7,$ $p<0.001$.
As with accuracy, we did not find any significant interactions with block number, but did find an interaction between condition and nameability, $ b = 0.28,~t(1372) = 2.3, p = 0.02$.
Although listeners were always a bit slower to respond for control tangrams than repeated tangrams, the gap was significantly bigger for low-nameability ones. 

\subsubsection{Stability of descriptions}

So far we have observed that dyads are able to communicate more efficiently and accurately over the course of a game, as a function of nameability.
However, these coarse metrics are agnostic to the actual linguistic content of descriptions, the vocabulary used by the speaker in order to describe targets. 
To examine whether interlocutors are truly converging on shared conventions (despite swapping roles each block), we calculated a measure called the Jaccard index, $J(W_i, W_{i+1}$), between the set of words $W_i$ used on successive blocks for tangrams in the \emph{repeated} condition.
The Jaccard index is a measure of set similarity defined as the size of the intersection over the size of the union, e.g. $$J(W_1, W_2) = |W_1 \cap W_2| / |W_1  \cup W_2|$$
To ensure effects were not driven by spurious changes in function words or pluralization, we lemmatized all descriptions and excluded stop words prior to computing sets.
The results of this analysis are shown in 
We found a steady increase in similarity across successive blocks (\autoref{si-fig: exp1_results}D), indicating that participants increasingly reused lexical choices from their partner in the previous block as shared conventions stabilized.\footnote{We found a similar effect for a directed variant of the Jaccard index: $J_{dir}(W_1, W_2)=|W_1 \cap W_2| / |W_2|$ normalizing by the size of the later description rather than the union. This supplemental analysis suggests that the observed changes in set similarity were not driven by non-stationarity in set size over time.}

\section{Pilot 2: Generalizing to unseen targets}\label{secA4}

\begin{figure}[!htb]
\centering
\includegraphics[width=0.6\textwidth,clip,trim=415 1485 295 13]{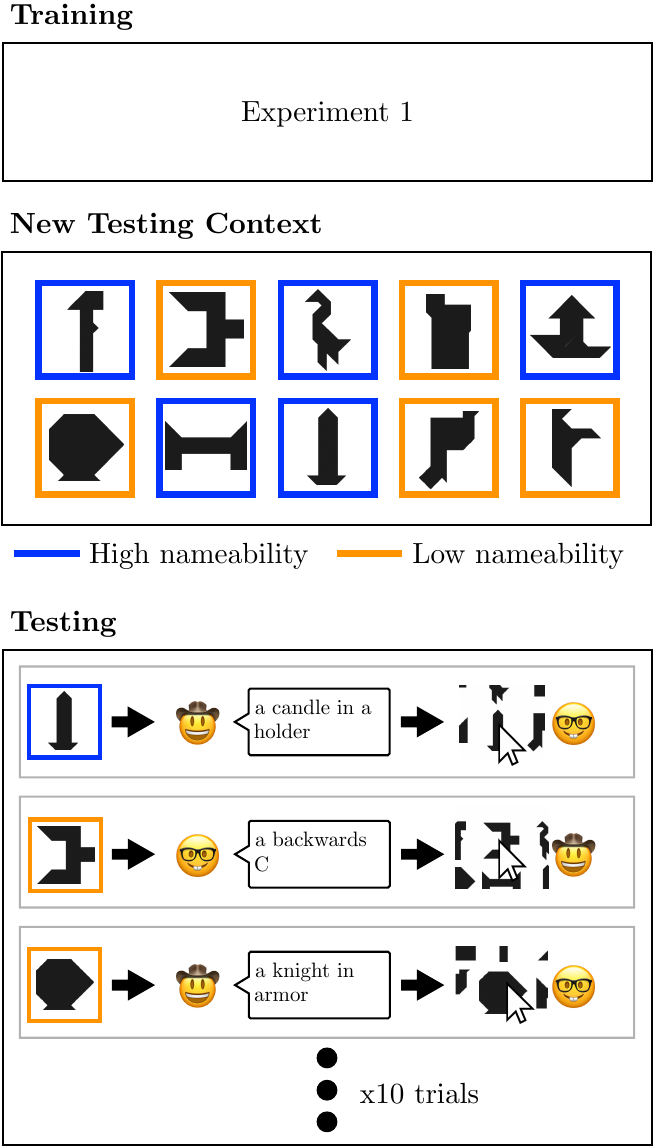}
\caption{\textbf{Pilot 2 design.} After participants completed the same procedure as Pilot 1 as a \emph{training phase}, they proceeded to a new \emph{test phase} where they played a reference game with an entirely new context of 10 tangrams.}
\label{si-fig: exp2}
\end{figure}

Our findings from Pilot 1 not only emphasize the added difficulty of communicating about low-nameability objects overall, but also hint at the additional difficulty of \emph{generalizing} newly acquired conventions to low-nameability objects. 
This effect was particularly strong for description length, where speakers were only willing to extend reduced descriptions to high-nameability control objects.
One explanation for the differential effect of nameability on control trials is that speakers became increasingly confident that their partner would share the same meaning for unseen tangrams only when they were expected to have high consensus \emph{a priori} and existing conventions were likely to be effective \citep{murthy2022shades}.
However, it is also possible that these control tangrams were simply more \emph{familiar} as they had appeared in context alongside the repeated tangrams prior to appearing as the target.
In Pilot 2, we introduce a stronger test of this hypothesis by measuring how speakers generalize to new contexts where all targets are entirely novel. 

\subsection{Participants}
We recruited 60 pairs of participants, 8 of whom were excluded based on the same criteria used in Pilot 1. 
Games lasted an average of 27 minutes and participants were paid \$5.50  (approximately \$11 per hour) with a performance bonus up to \$1.20.

\subsection{Stimuli, design and procedure}
We used the same sets of high- and low-nameability stimuli from Pilot 1, and the procedure was a direct replication and extension. 
The first phase of the Pilot (the \emph{training} phase) was an exact replication of the within-dyad $2 \times 2$ design used in Pilot 1.
The second phase (the \emph{test} phase) was new.
A 6th block was appended, containing 10 additional trials (\autoref{si-fig: exp2}).
Critically, these test trials used a completely non-overlapping context with 10 new targets presented in a randomized sequence. 
Each tangram in the new context was given a single trial. 
Speaker and listener roles were swapped between every trial.

\subsection{Results}

We evaluated the performance of a new context by examining two metrics: accuracy and verbosity (we omit the Pilot 2 response time analysis due to space constraints).

\begin{figure}[t]
\centering
\includegraphics[width=0.7\textwidth]{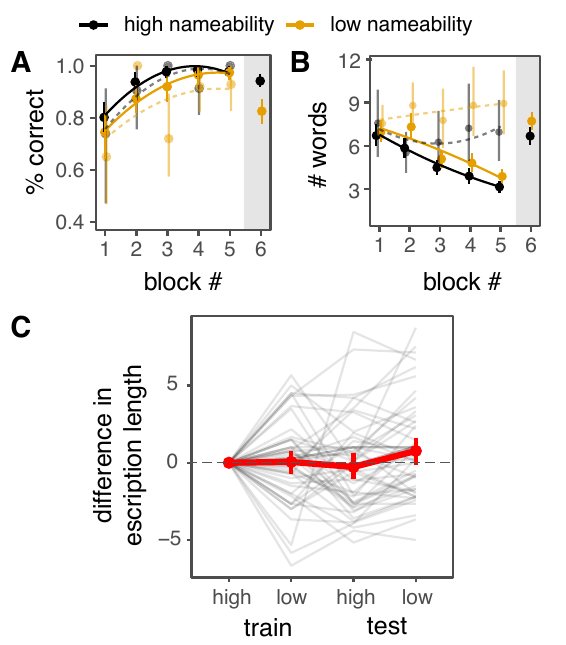}
\caption{\textbf{Pilot 2 results}. (A) Accuracy and (B) description length for the train phase (blocks 1-5) and test phase (block 6). Transparent dashed lines are control conditions. (C) Within-game differences in description length relative to the first block of train for the first train and test blocks, for high- and low-nameability targets. Error bars are bootstrapped 95\% CIs.} 
\label{si-fig: exp2_results}
\end{figure}

\subsubsection{Accuracy}

In addition to replicating the nameability effects examined in Pilot 1 (see \autoref{si-fig: exp2_results}A), the primary analysis of interest is a direct comparison against the initial block of the train phase (block 1) and the initial block of the test phase (block 6).
In both cases, it is the speaker's first time referring to all targets in context.
Thus, any differences in the test phase can be attributed to some generalizable learning taking place over the training block.
We ran a mixed-effects logistic regression model predicting accuracy only for these two blocks, including fixed effects of phase (train vs. test) and nameability (low vs. high), as well as random intercepts and slopes at the dyad-level.
We found a significant interaction, $b=-0.31, z=2.25, p=0.024$, indicating that there was greater improvement from the beginning of train to test for high-nameability tangrams than low-nameability ones.

\subsubsection{Description length}
Average description lengths are shown for the training phase and test phase in \autoref{si-fig: exp2_results}B. 
We again replicated the nameability effects found in Pilot 1, and focus here on the direct comparison between the first block of training and the first block fo test. 
We ran a mixed-effects linear regression model predicting description length, including fixed effects of phase (train vs. test) and nameability (low vs. high) with random intercepts and slopes at the dyad-level.
In addition to a significant main effect of nameability at the beginning of both train and test, $b=0.32, t(112)=-2.181, p=  0.03$, we found a marginal interaction, $b=0.25, t(629)=1.8, p = 0.069$. 
In a Bayesian mixed-effects model with full random effects, we obtained a 95\% credible interval of $[-0.05, 0.54]$. 
In other words, there was  a small but meaningful gap in description length between high and low-nameability tangrams in the test phase, while no such gap was observed in the first block (see \autoref{si-fig: exp2_results}C for a finer-grained visualization controlling for individual differences in overall verbosity). 
Put together, these effects suggest that participants were better able to anticipate in the test phase which tangrams would be harder and adjust their description length accordingly.

\section{Experiment interface}\label{secA5}
\begin{figure}[t]
\centering
\includegraphics[width=\textwidth]{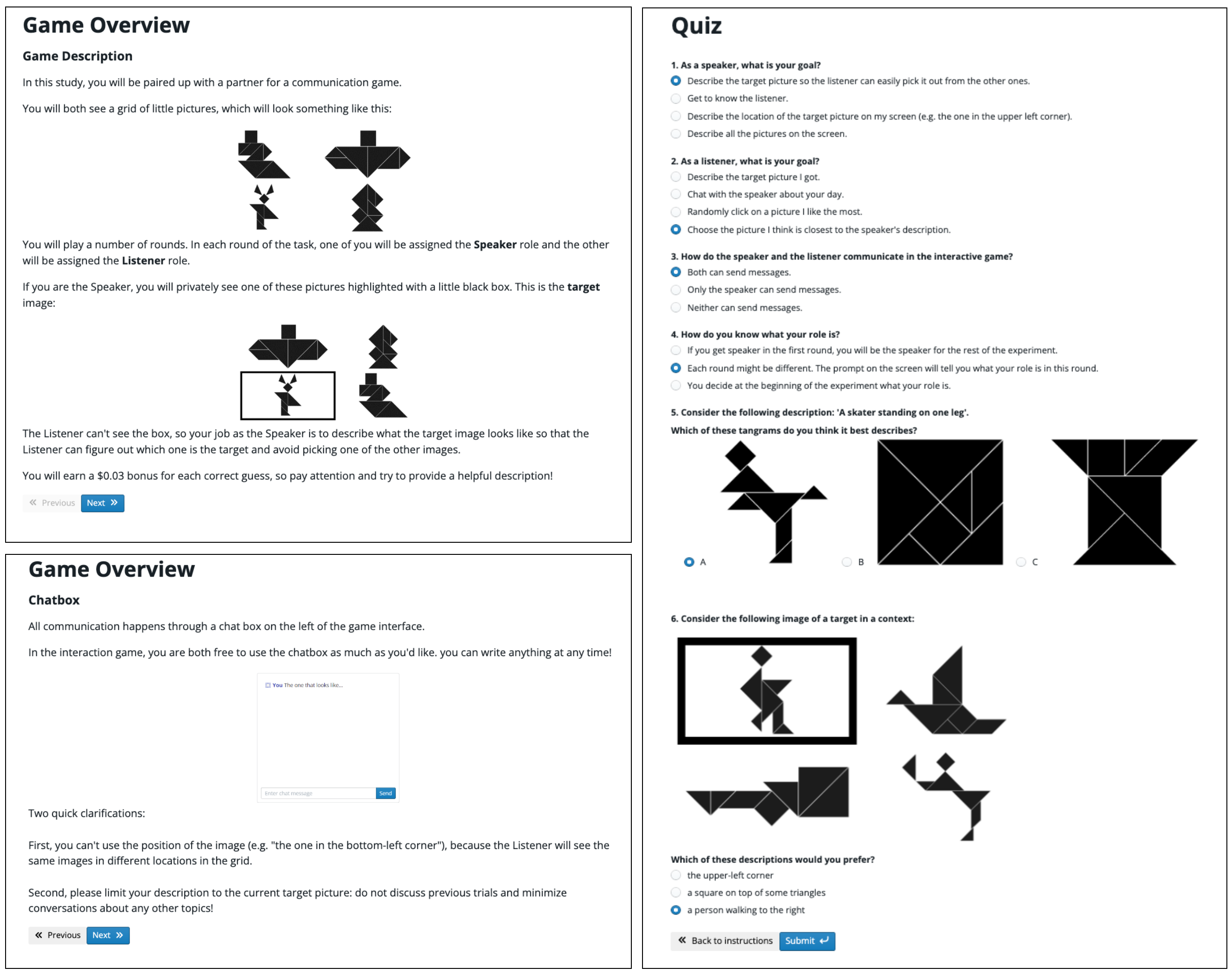}
\caption{Experiment introduction and qualification quiz.}
\label{si-fig: exp-intro}
\end{figure}

\begin{figure}
\centering
\includegraphics[width=\textwidth]{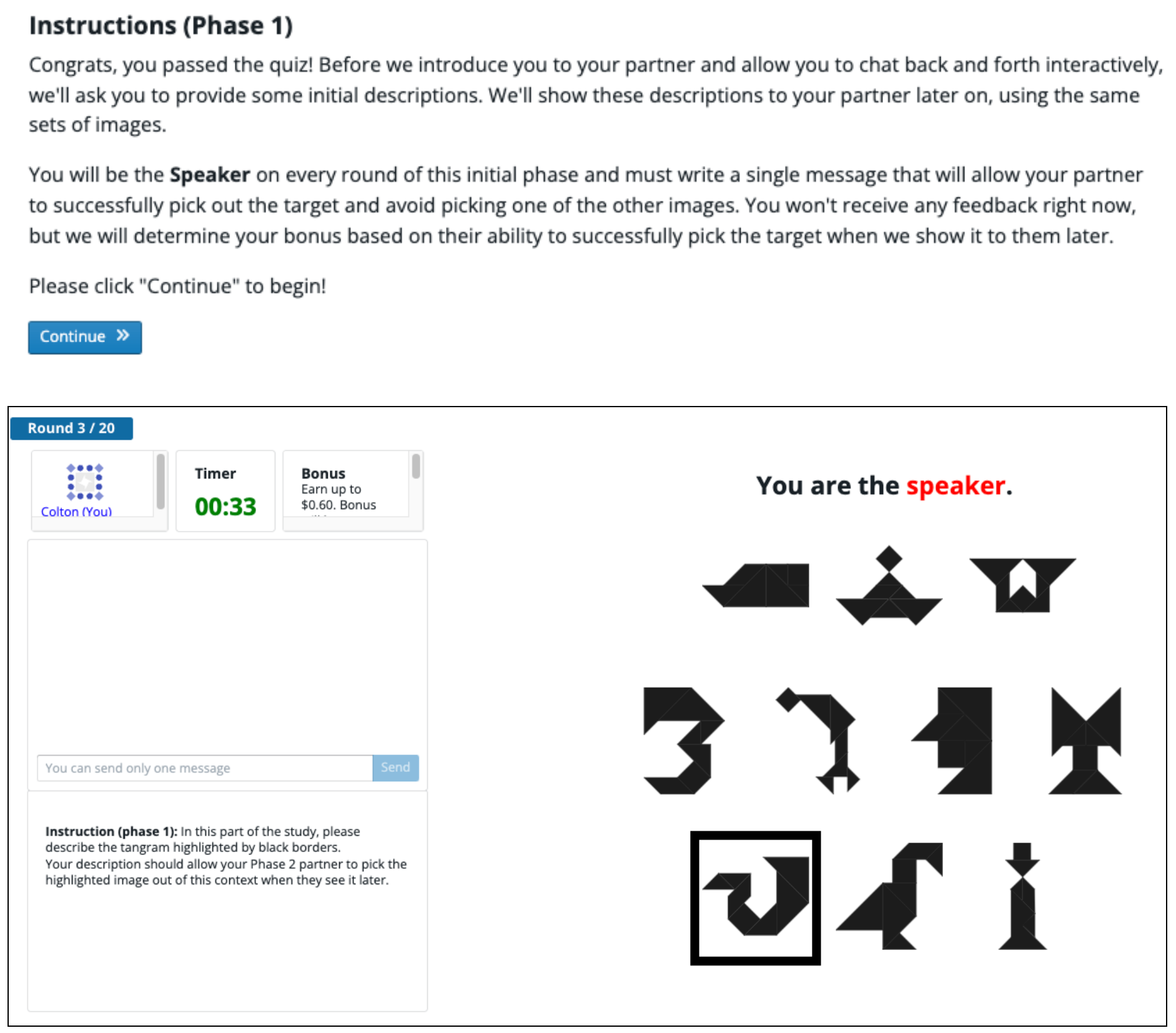}
\caption{Pre-interactive phase instructions and task interface.}
\label{si-fig: exp-pre}
\end{figure}

\begin{figure}
\centering
\includegraphics[width=0.8\textwidth]{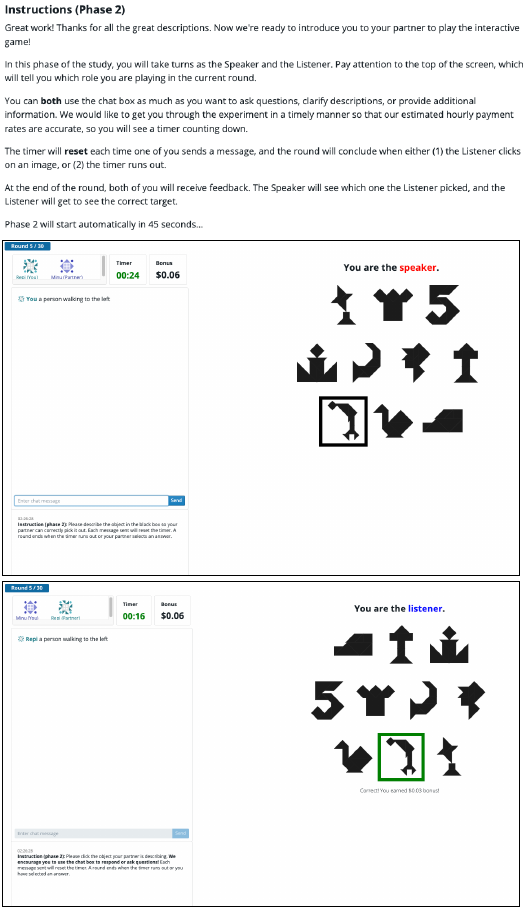}
\caption{Interactive phase instructions and task interface.}
\label{si-fig: exp-interactive}
\end{figure}

\begin{figure}
\centering
\includegraphics[width=0.9\textwidth]{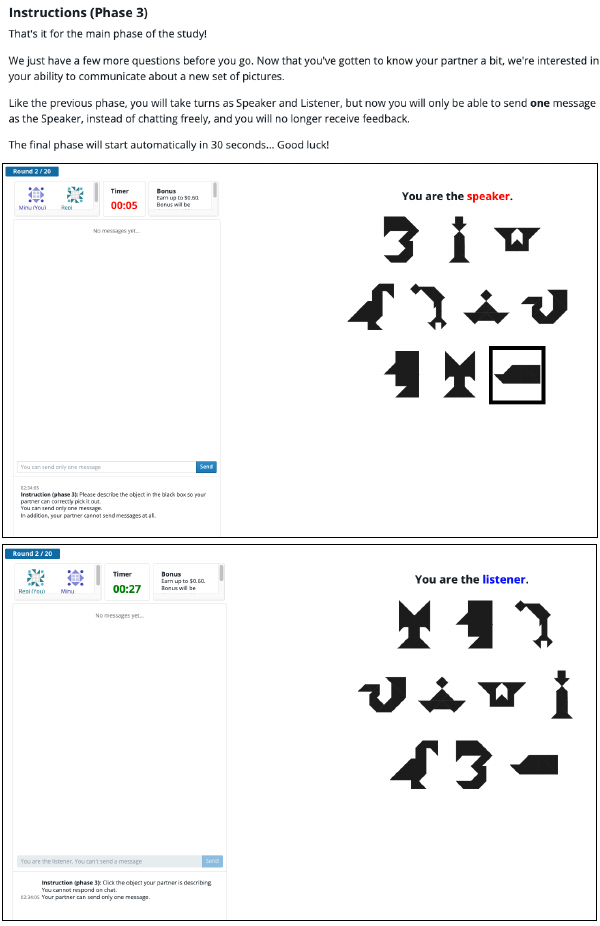}
\caption{Post-interactive phase instructions and task interface.}
\label{si-fig: exp-post}
\end{figure}

\autoref{si-fig: exp-intro} - \autoref{si-fig: exp-post} are screenshots of the interface for the experiment in the main text.

\end{appendices}

\end{document}